\documentclass[pdflatex,sn-mathphys-ay,iicol]{sn-jnl}

\usepackage{graphicx}%
\usepackage{tabularx}
\usepackage{multirow}%
\usepackage{amsmath,amssymb,amsfonts}%
\usepackage{amsthm}%
\usepackage{mathrsfs}%
\usepackage[title]{appendix}%
\usepackage{xcolor}%
\usepackage[table]{xcolor}
\usepackage{pifont}
\usepackage{makecell}
\usepackage{textcomp}%
\usepackage{manyfoot}%
\usepackage{booktabs}%
\usepackage{subcaption}
\usepackage{algorithm}%
\usepackage{algorithmicx}%
\usepackage{algpseudocode}%
\usepackage{listings}%
\usepackage{cleveref}
\newcommand{\p}[4]{%
    \ifx#1\empty
        #3
    \else
        \cellcolor{#1}#3
    \fi
    \quad
    \ifx#2\empty
        #4
    \else
        \cellcolor{#2}#4
    \fi
}

\usepackage{stackengine} 

\definecolor{b}{rgb}{1, 1,1}
\definecolor{fst}{rgb}{0.628, 0.565, 0.781}
\definecolor{sec}{rgb}{0.753, 0.741, 0.910}
\definecolor{thd}{rgb}{0.922, 0.904, 1}

\definecolor{cap}{rgb}{0.408, 0.267, 0.545}

\newcommand{\cmark}{\textcolor{green}{\ding{51}}} 
\newcommand{\xmark}{\textcolor{red}{\ding{55}}}   
\newcommand{\goldstar}{\textcolor{yellow}{\ding{72}}}

\newcolumntype{P}[1]{>{\raggedright\arraybackslash}p{#1}}

\newcolumntype{P}[1]{>{\raggedright\arraybackslash}p{#1}}

\newcommand{\BigColBegin}{%
\setlength{\tabcolsep}{3pt} 
\begin{tabular*}{\textwidth}{@{\extracolsep{\fill}}
c@{\hspace{3pt}}|@{\hspace{3pt}} 
@{}p{0.4\textwidth}@{}          
@{\hspace{4pt}}|@{\hspace{4pt}}  
c c
@{\hspace{4pt}}|@{\hspace{4pt}}  
c c
@{\hspace{4pt}}|@{\hspace{4pt}}  
c c@{}}
\hline
 & \textbf{Methods} &
\multicolumn{2}{c@{\hspace{4pt}}|@{\hspace{4pt}}}{\textbf{PSNR}$\uparrow$} &
\multicolumn{2}{c@{\hspace{4pt}}|@{\hspace{4pt}}}{\textbf{SSIM}$\uparrow$} &
\multicolumn{2}{c}{\textbf{LPIPS}$\downarrow$} \\
 & &
\small Train & \small NVS &
\small Train & \small NVS &
\small Train & \small NVS \\
\hline
}
\newcommand{\BigColEnd}{\hline\end{tabular*}}

\newcommand{\TypeFirst}[9]{%
\multirow{#2}{*}{\rotatebox{90}{\scriptsize\textbf{#1}}} &
#3 & #4 & #5 & #6 & #7 & #8 & #9 \\
}
\newcommand{\TypeNext}[7]{%
& #1 & #2 & #3 & #4 & #5 & #6 & #7 \\
}

\newcommand{\TypeSep}{%
  \noalign{\vskip 5pt}%
  \hline
  \noalign{\vskip 5pt}%
}

\theoremstyle{thmstyleone}%
\theoremstyle{thmstyletwo}%

\theoremstyle{thmstylethree}%

\begin{document}

\title[Article Title]{RealX3D: A Physically-Degraded 3D Benchmark for Multi-view Visual Restoration and Reconstruction}


\author*[1]{\fnm{Shuhong} \sur{Liu}}\email{s-liu@mi.t.u-tokyo.ac.jp}
\equalcont{These authors contributed equally to this work.}

\author[1]{\fnm{Chenyu} \sur{Bao}}\email{c-bao@mi.t.u-tokyo.ac.jp}
\equalcont{These authors contributed equally to this work.}

\author[1]{\fnm{Ziteng} \sur{Cui}}\email{cui@mi.t.u-tokyo.ac.jp}

\author[2]{\fnm{Yun} \sur{Liu}}\email{yunliu@nii.ac.jp}

\author[1]{\fnm{Xuangeng} \sur{Chu}}\email{xuangeng.chu@mi.t.u-tokyo.ac.jp}

\author[3]{\fnm{Lin} \sur{Gu}}\email{lingu.edu@gmail.com}

\author[4]{\fnm{Marcos V.} \sur{Conde}}\email{marcos.conde@uni-wuerzburg.de}

\author[1]{\fnm{Ryo} \sur{Umagami}}\email{umagami@mi.t.u-tokyo.ac.jp}

\author[1]{\fnm{Tomohiro} \sur{Hashimoto}}\email{hashimoto@mi.t.u-tokyo.ac.jp}

\author[1]{\fnm{Zijian} \sur{Hu}}\email{zijian.hu@mi.t.u-tokyo.ac.jp}

\author[1]{\fnm{Tianhan} \sur{Xu}}\email{tianhan.xu@mi.t.u-tokyo.ac.jp}

\author[1]{\fnm{Yuan} \sur{Gan}}\email{y-gan@mi.t.u-tokyo.ac.jp}

\author[1]{\fnm{Yusuke} \sur{Kurose}}\email{kurose@mi.t.u-tokyo.ac.jp}

\author[1,5]{\fnm{Tatsuya} \sur{Harada}}\email{harada@mi.t.u-tokyo.ac.jp}

\affil[1]{\orgname{The University of Tokyo}, \orgaddress{\street{4 Chome-6-1 Komaba}, \city{Meguro}, \postcode{153-8904}, \state{Tokyo}, \country{Japan}}}

\affil[2]{\orgname{National Institute of Informatics}, \orgaddress{\street{2 Chome-1-2 Hitotsubashi}, \city{Chiyoda}, \postcode{101-8430}, \state{Tokyo}, \country{Japan}}}

\affil[3]{\orgname{Tohoku University}, \orgaddress{\street{41 Kawauchi}, \city{Aoba}, \postcode{980-8576}, \state{Sendai}, \country{Japan}}}

\affil[4]{\orgname{University of Würzburg}, \orgaddress{\street{Sanderring 2}, \city{City}, \postcode{97070}, \state{Würzburg}, \country{Germany}}}

\affil[5]{\orgname{RIKEN AIP}, \orgaddress{\street{1-4-1 Nihonbashi}, \city{Chuo}, \postcode{103-0027}, \state{Tokyo}, \country{Japan}}}


\abstract{Reliable 3D reconstruction is a prerequisite for robotics, embodied AI, and immersive AR/VR applications; however, real-world observations frequently depart from clean imaging assumptions due to illumination changes, participating media, occlusions, and blur that break multi-view consistency and destabilize pose estimation, which leaves a clear gap between performance on curated benchmarks and behavior in practical deployments. To address this gap, we introduce RealX3D, a real capture benchmark for multi-view restoration and reconstruction under real-world degradations, organized into four families spanning nine controlled settings that include motion blur, defocus blur, low-light, view-varying exposure, smoke, dynamic occlusion, and reflection. RealX3D is collected using a unified acquisition protocol that enables recapturing the same camera trajectories to obtain pixel-aligned low-quality and reference ground-truth image pairs. Each scene also provides per-view RAW measurements to preserve high dynamic range linear sensor signals. To support geometry-grounded evaluation beyond image photometric fidelity, we capture dense laser scan geometry for every scene and derive world-scale measures such as point clouds, meshes, and metric depth, allowing comprehensive assessment of pose, depth, and surface reconstruction alongside photometric restoration quality. The benchmark contains 55 scenes recorded at high resolution with diverse real-world degradation patterns. We benchmark a broad set of optimization-based and feed-forward methods using both image metrics and geometry metrics, and the results reveal substantial robustness gaps across degradations in adverse conditions. Overall, RealX3D provides a rigorous benchmark that moves beyond synthetic data and establishes a standardized foundation for developing degradation-robust 3D reconstruction systems.}

\keywords{3D Reconstruction, Multi-view Visual Restoration, Physical Degradation}



\maketitle

\begin{figure*}[!ht]
\centering
\includegraphics[width=1\textwidth]{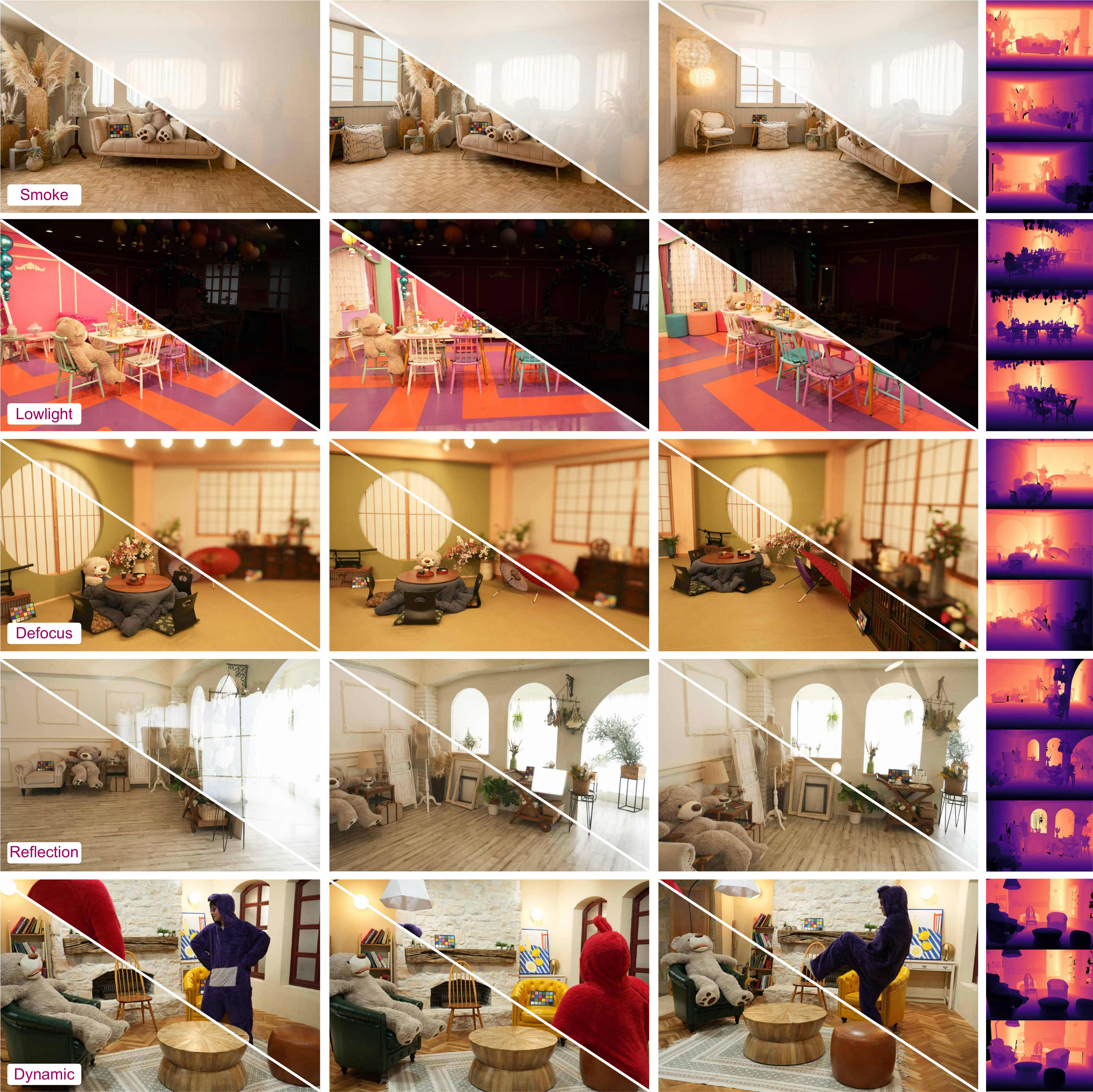}
\caption{RealX3D is a real-capture benchmark for 3D reconstruction under real-world degradations. It spans four general types across nine levels, including motion and defocus blur, lowlight, varying exposure, smoke, dynamic occlusion, and reflection. Each scene provides pixel-aligned low-quality and reference views, RAW data, and dense laser-scan geometry for comprehensive evaluation.}
\label{fig:teaser}
\end{figure*}

\section{Introduction}
\label{sec:intro}

3D reconstruction and novel view synthesis (NVS) have become foundational components of embodied AI and spatially grounded systems. Robots and autonomous vehicles rely on accurate 3D scene representations for navigation, planning, and interaction \citep{xie2022neural,zhou2024mod,li2024sgs,mascaro2025scene,zhao2025survey}. Likewise, AR and VR applications require stable geometry and view-consistent rendering to anchor virtual content reliably in the physical world \citep{itoh2021towards,cao2023mobile}. In practice, however, real-world image capture rarely satisfies ideal imaging assumptions. Low illumination, reflections, smoke, occlusions, and blur frequently corrupt observations, creating a persistent gap between controlled laboratory settings and in-the-wild deployment \citep{kwon2025r3evision,lidense2025,liu2025mg,zhao2025resilient}.

Recent advances in neural scene representations, such as Neural Radiance Fields (NeRF) \citep{mildenhall2021nerf} and 3D Gaussian Splatting (3DGS) \citep{kerbl20233d}, have substantially improved reconstruction fidelity and rendering quality under favorable conditions. Nevertheless, robustness remains fragile when inputs violate spatio-temporal consistency due to real-world degradations. Pose estimation further exacerbates this challenge. Most reconstruction pipelines are initialized using Structure-from-Motion (SfM) \citep{schonberger2016structure}, which estimates camera poses and sparse geometry from feature correspondences. Under adverse imaging conditions, degraded feature detection and matching often lead to biased pose estimates or complete pose failure. As a result, robust 3D reconstruction must simultaneously contend with appearance corruption and pose unreliability induced by real-world capture.

To mitigate these issues, recent work has increasingly incorporated image formation models and robustness mechanisms into the reconstruction pipeline. Representative directions include jointly modeling degradation and scene representation \citep{martin2021nerf,ma2022deblur,oh2024deblurgs}, introducing physically motivated rendering models and consistency constraints \citep{ramazzina2023scatternerf,levy2023seathru,liu2025i2nerf}, leveraging learned priors to stabilize optimization under challenging conditions \citep{ren2024nerf,kulhanek2024wildgaussians,sabour2025spotlesssplats}, and coupling multi-view restoration modules with 3D parameter updates \citep{zhang2024gaussian,liu2025deraings,choi2025exploiting}. Complementary lines of research focus on alleviating the pose bottleneck through robust correspondence strategies, uncertainty-aware pose refinement, or learning-based pose correction integrated into reconstruction pipelines \citep{lee2023exblurf,wang2023bad,zhao2024bad,lu2025bard}.

Despite this methodological progress, benchmarking and evaluation have not kept pace. Many existing datasets rely on synthetic degradations that inadequately reflect real sensor pipelines and physical image formation processes \citep{chen2024dehazenerf,sun2024dyblurf,liu2025deraings,ma2025dehazegs}. Real-capture benchmarks, by contrast, are typically tailored to specific degradation types and acquisition setups \citep{wang2023lighting,sabour2023robustnerf,cui2024aleth}. They often offer limited viewpoint coverage or scene diversity, or capture degraded and clean sequences along different trajectories. Such mismatches break pixel-level correspondence and complicate direct comparisons across methods \citep{snavely2006photo,zhang2021learning,ma2022deblur,lee2023exblurf,ren2024nerf}. For 3D reconstruction and NVS, rigorous evaluation benefits from pixel-aligned low-quality (LQ) observations and clean ground-truth (GT) images captured from identical viewpoints. Moreover, current evaluations are frequently restricted to photometric fidelity and lack reliable geometric references, which are essential for assessing geometric accuracy and view-consistent rendering.

To address these limitations, we introduce RealX3D, a high-resolution benchmark for 3D reconstruction and NVS under real-world degradations, captured within a unified acquisition protocol. RealX3D organizes real degradations into four major categories: illumination, scattering, occlusion, and blurring. Each category is instantiated through controlled yet physically realistic capture procedures. Crucially, RealX3D provides pixel-aligned low-quality and ground-truth image pairs wherever physically feasible by re-capturing identical trajectories using a high-precision rail-based camera dolly system. Beyond standard RGB (sRGB) imagery, we store per-view RAW measurements to preserve richer linear signals under severe degradations and to support RAW-space reconstruction and evaluation. Each scene is further captured with high-end laser scanning, enabling dense geometry acquisition, metric depth generation, and reliable geometric evaluation. \Cref{fig:teaser} visualizes three representative LQ and GT training pairs from selected subcategories in RealX3D, together with the corresponding metric depth maps. To prevent pose instability from confounding reconstruction performance, we estimate camera poses on the corresponding reference views using calibrated intrinsics and apply these poses consistently to both low-quality and reference sequences. We further register SfM reconstructions with laser-scan coordinates to achieve accurate real-world alignment.

The contributions of our work are summarized as:
\begin{itemize}
    \item We introduce a real-world 3D restoration and reconstruction dataset that contains diverse degradation types of illumination, scattering, occlusion, and blurring.
    \item We develop a physically-degraded data acquisition pipeline that offers pixel-aligned LQ/GT pairs, metric depth, scanned point clouds, and extracted meshes for comprehensive evaluation.
    \item We conduct comprehensive studies and evaluations for existing methods on each specific task using our proposed benchmark dataset, and reveal substantial performance gaps that highlight the challenges posed by real-world degradations.
\end{itemize}

\section{Related Work}

We begin with a brief overview of recent methods and datasets organized by each specific degradation type. For reconstruction methods, we distinguish between optimization-based approaches, such as NeRF- and 3DGS–based pipelines that typically rely on SfM initialization, and recent feedforward foundation models that simultaneously estimate camera pose and scene geometry in a single forward pass.

\subsection{Robust 3D Optimization and Reconstruction Approches}

Recent optimization-based reconstruction methods for degraded imagery are driven by a mismatch between the idealized assumptions of reconstruction methods and the realities of in-the-wild capture, motivating models that couple reconstruction with image-formation physics or robust priors.

\subsubsection{Deblurring Method}

Finite exposure integrates radiance over time, and relative camera or scene motion converts this temporal integration into spatially varying, depth-dependent blur that disrupts correspondence and multi-view consistency. Deblur-NeRF \citep{ma2022deblur} initiates blur-aware neural reconstruction by embedding a differentiable blur formation model into NeRF, jointly estimating latent blur parameters and a sharp radiance field. Subsequent NeRF-based extensions advance blur-aware reconstruction along three parallel directions: physics-inspired constraints and depth-aware blur modeling to stabilize kernel estimation \citep{lee2023dp}; exposure-time camera trajectory optimization in a bundle-adjustment-like manner to improve robustness under inaccurate pose initialization \citep{wang2023bad,lee2023exblurf}; and practical training schemes with progressive deblurring as well as extensions to video and dynamic settings \citep{peng2023pdrf,sun2024dyblurf,luo2024dynamic}. In parallel, blur-aware modeling has been adapted to explicit 3DGS pipelines \citep{zhao2024bad,oh2024deblurgs,lee2024deblurring,lu2025bard,wang2025dof}, where blur is handled through exposure-time motion estimation, blur-aware rendering of splats, or parameterizations that modulate Gaussian covariances, while preserving the efficiency of rasterization-based rendering. Finally, hybrid approaches \citep{wang2024mpnerf,li2024rustnerf,park2024towards,choi2025exploiting,bui2025mobgs} incorporate deblurring priors from image restoration networks into the reconstruction loop, using learned deblurring as a regularizer to alleviate the ill-posedness of recovering sharp geometry and appearance from heavily blurred multi-view observations.

\subsubsection{Illumination-Robust Method}

In an extreme low-light scenario, photon scarcity amplifies noise, while ISP nonlinearities distort radiometric consistency and invalidate brightness constancy. RAW-space methods address this by reconstructing directly from linear sensor measurements, where exposure can be modeled explicitly. RawNeRF  \citep{mildenhall2022nerf} optimizes NeRF on noisy RAW inputs and recovers an HDR radiance field that can be re-exposed and tone-mapped at render time. Recent RAW-domain pipelines \citep{li2024chaos,wang2025bright} analyze noise sensitivity and introduce stabilization strategies, including sensor-response-inspired constraints and RAW-consistent color adaptation, to prevent degeneration and improve denoising and exposure normalization. Since low-light capture often co-occurs with saturation and clipping, HDR-oriented methods explicitly parameterize exposure and tone mapping to recover radiance beyond LDR limits. HDR-NeRF \citep{huang2022hdr} learns an HDR field from multi-exposure supervision via exposure-conditioned rendering, and rasterization-based methods \citep{cai2024hdr,jin2024lighting} bring similar principles to 3DGS for more efficient optimization. Recent variants refining tone mapping and exposure modeling for superior NVS rendering \citep{wang2024bilateral,liu2025gausshdr,niemeyer2025learning}.

On the other hand, sRGB-space approaches operate on camera-finished images and focus on learning view-consistent enhancement jointly with reconstruction. LLNeRF \citep{wang2023lighting} integrates low-light decomposition and enhancement into NeRF optimization to avoid per-image, view-inconsistent preprocessing, while Aleth-NeRF \citep{cui2024aleth}  and subsequent studies \citep{zhang2024ambient,xu2025physical} model adverse illumination through adaptive transmittance in volume rendering. Recent 3DGS-based approaches \citep{qu2024lush,cui2025luminance,zhou2025lita,sun2025ll,li2025robust} introduce view-adaptive photometric transforms and illumination-aware priors to better tolerate exposure shifts and color changes during splat optimization. In addition to illumination, physically grounded media-interaction models provide a broader unifying perspective that naturally covers low-light conditions together with absorption and scattering effects \citep{liu2025i2nerf}.

\subsubsection{Occlusion-Removal Method}

Casual captures contain transient occluders, distracting objects, reflections, and appearance changes that violate multi-view photometric consistency. NeRF-in-the-Wild \citep{martin2021nerf} explicitly factorizes a static field and transient components via per-image appearance and transient embeddings with uncertainty-weighted training. Subsequent methods further improve robustness to occluders by incorporating visibility-guided anti-occlusion mechanisms, cross-ray interactions with occlusion-aware objectives, and robust estimation that down-weights distractors without explicit semantic masks \citep{chen2022hallucinated,yang2023cross,sabour2023robustnerf,ren2024nerf}.

Robustness has also been translated to 3DGS. GS-W \citep{zhang2024gaussian} introduces per-Gaussian intrinsic and dynamic appearance features together with visibility modeling to reduce the influence of transient occluders in unconstrained collections, and WildGaussians \citep{kulhanek2024wildgaussians} incorporates robust features with uncertainty-driven masking to stabilize splat optimization under occlusions. SpotLessSplats \citep{sabour2025spotlesssplats} leverages strong pretrained features for clustering and robust masking to ignore transient distractors, while DeGauss \citep{wang2025degauss} and HybridGS \citep{lin2025hybridgs} explicitly decompose dynamic and static components using separate Gaussian sets or joint 2D and 3D masking to obtain distractor-free reconstructions. DeSplat \citep{wang2025desplat} achieves explicit separation by jointly optimizing static Gaussians and per-view distractor Gaussians using only splatting-based rendering and photometric supervision, and RogSplat \citep{kong2025rogsplat} adds generative priors to detect unreliable regions and refine occluded content during optimization. 

\subsubsection{Scattering-Aware Method}

In haze or smoke, scattering and absorption reduce transmittance and introduce additive path radiance, so reconstruction methods often embed radiative-transfer terms to decouple direct scene radiance from scattered components under multi-view constraints. A representative formulation integrates physically grounded transmittance and in-scattering into neural rendering for joint reconstruction and dehazing \citep{ramazzina2023scatternerf}, and subsequent work improves robustness through additional physical priors \citep{li2023dehazing,chen2024dehazenerf,zhang2025decoupling} and by transferring the same formation model to 3DGS pipelines \citep{ma2025dehazegs}.

Underwater scenes further amplify these effects because attenuation is strongly wavelength dependent and backscatter grows quickly with range, making color restoration and geometry recovery tightly coupled. SeaThru-NeRF \citep{levy2023seathru} provides a canonical physics-guided approach by explicitly modeling underwater image formation to recover a cleaner radiance field suitable for novel-view rendering. Follow-up methods extend this idea to handle more complex real captures and improve efficiency, including medium-aware splatting formulations \citep{li2025watersplatting,yang2025seasplat}, stronger priors for separating medium effects from scene appearance \citep{tang2024neural,gough2025aquanerf,wu2025plenodium,guo2025neuropump}, and unified media-interaction models that treat underwater and illumination within a single framework \citep{liu2025i2nerf}.

\subsubsection{Method for Adverse-Weather}

Rain and snow corrupt multi-view captures through a mixture of atmospheric particles and lens-attached droplets, producing streak-like distortions, refraction, and intermittent occlusions that violate view consistency and may be absorbed into the reconstructed geometry. NeRF-based pipelines \citep{li2024derainnerf,lyu2024rainyscape,li2024derainnerf} therefore model weather effects explicitly, for example by predicting droplet visibility or decomposing a transient weather layer, so that radiance field optimization can focus on the underlying static scene. More recent Gaussian-splatting frameworks emphasize separating dense particle artifacts from sparse lens occlusions and using mask-aware optimization to prevent weather artifacts from being reconstructed as persistent structure \citep{liu2025deraings,qian2025weathergs}. Pipeline-level analyses further show that precipitation can also break SfM pose estimation and point-cloud initialization, motivating joint designs that stabilize both preprocessing and reconstruction under unconstrained rainy inputs \citep{yang2025rethinking}. Complementary to method design, controllable simulation pipelines provide systematic stress tests for adverse-weather reconstruction by enabling view-consistent rain and snow synthesis on Gaussian scenes, typically using physics-inspired particle models or diffusion- and score-distillation-based generation to animate weather dynamics while preserving underlying geometry \citep{dai2025rainygs,qian2025weatheredit,fiebelman2025letitsnow}.

\begin{sidewaystable*}[!htp]
    \centering
    \caption{Comparison of existing degraded 3D datasets. RealX3D surpasses existing datasets in diversity and resolution, and further offers RAW sensor data that preserves richer signals under severe degradations, alongside high-end laser scans for precise geometry capture. \textbf{Img/s} denotes the average number of images per scene. \textbf{Scan} indicates the availability of scanned point clouds. \textbf{Pair-GT} denotes paired ground truth, where a dataset provides pixel-aligned clean images for either synthetic or real-world data. \textbf{Depth} indicates the availability of real-world depth measurements. \textbf{NVS} denotes held-out test views for novel-view synthesis. \textbf{Raw} indicates the availability of raw sensor images. $^*$ indicates resolution differs across scenes; we report the maximum resolution.}
    \small
    \setlength{\tabcolsep}{4pt}
    \setlength{\extrarowheight}{2pt}
    \begin{tabularx}{\textwidth}{lllccccccccc}
    \toprule
    Dataset & Venue & Degrad. Type & Method & Total Scene & Resolution & Img/s & Scan & Pair-GT  & Depth & NVS & Raw \\ 
    \midrule
    Deblur-NeRF \citep{ma2022deblur} &  CVPR22  & Motion/Defocus & Real & 25 & 2400x1600$^*$ &39 & \xmark & \cmark & \xmark & \xmark & \xmark\\ 
    ExBluRF \citep{lee2023exblurf} & CVPR23   & Motion & Real & 8 & 800x540 & 30 & \xmark &\cmark  & \xmark & \xmark& \xmark\\
    DyBluRF \citep{sun2024dyblurf} &  CVPR24  & Motion & Syn & 6 &1280x720 & 24 & \xmark & \cmark & \xmark & \xmark & \xmark\\
    D2RF \citep{luo2024dynamic} & ECCV24& Defocus& Syn & 8&1880x800& 23&\xmark&\cmark&\cmark&\xmark&\xmark\\
    BARD-GS \citep{lu2025bard} & CVPR25 & Motion & Syn & 12 & 960x540 & 74 & \xmark &\cmark &\xmark & \cmark &\xmark \\
    BlurRF \citep{choi2025exploiting} & CVPR25 & Motion/Defocus & Syn/Real & 75/5 & 600x400 & 29 &\xmark &\cmark/\xmark & \xmark& \cmark& \xmark \\
    BlurryIPhone \citep{bui2025moblurf} & TPAMI25 & Motion & Syn & 7 & 360x480 & 365 & \xmark & \cmark & \cmark & \cmark & \xmark  \\
    \midrule
    Phototourism \citep{snavely2006photo} & IJCV20 & Occlusion & Real & 25 & - & 150 & \xmark & \xmark & \xmark & \xmark & \xmark \\
    D2NeRF \citep{wu2022d} & NIPS22 & Occlusion & Syn/Real & 5/10 & 512x512 & 200 & \xmark & \xmark & \xmark & \cmark & \xmark \\
    RobustNeRF \citep{sabour2023robustnerf} & CVPR23 & Occlusion & Syn/Real & 3/4 & 4032x3024 & 110 & \xmark & \cmark/\xmark & \xmark & \cmark & \xmark \\
    NeRF-Go \citep{ren2024nerf} & CVPR24 & Occlusion & Real & 12 & 4032x3024$^*$ & 180 & \xmark & \xmark & \xmark & \cmark & \xmark \\
    \midrule
    RawNeRF \citep{mildenhall2022nerf} & CVPR22 & Lowlight & Real & 14 & 4032x3024 & 56 & \xmark & \xmark & \xmark & \xmark & \cmark \\
    LLNeRF \citep{wang2023lighting} & ICCV23 & Lowlight & Real & 16 & 1156x858 & 25 & \xmark & \xmark & \xmark & \cmark & \xmark \\
    AlethNeRF \citep{cui2024aleth} & AAAI24 & Lowlight & Real & 5 & 500x375 & 36 & \xmark & \cmark & \xmark & \xmark & \xmark \\
    LuSh-NeRF \citep{qu2024lush} & NIPS24 & Lowlight & Syn/Real & 5/5 & 1120x640 & 22 & \xmark & \cmark/\xmark & \xmark & \xmark & \xmark \\
    \midrule
    REVIDE \citep{zhang2021learning} & CVPR21 & Haze Scattering & Real & 48 & 2708x1800 & 320 & \xmark & \cmark & \xmark & \xmark & \xmark \\
    SeaThruNeRF \citep{levy2023seathru} & CVPR23 & Smoke/Haze & Syn & 1 & 1008x756 & 20 & \xmark & \cmark & \xmark & \xmark & \xmark \\
    NeRF-dehaze \citep{jin2024reliable} & OptEx24 & Smoke/Haze & Real & 5 & 1920x1080$^*$ & 55 &  \xmark & \xmark & \xmark & \xmark & \xmark \\
    \midrule
    \textbf{RealX3D} &  & All Above \goldstar & Real & 55 & 7008x4672 & 30 & \cmark & \cmark & \cmark & \cmark & \cmark \\
    \bottomrule
    \end{tabularx}
    \label{tab:exist_dataset}
\end{sidewaystable*}

\subsection{Feedforward Geometry Foundation Model}

Recent feedforward geometry foundation models reduce dependence on SfM initialization by directly predicting camera and dense 3D attributes in a single forward pass. DUSt3R \citep{wang2024dust3r} popularizes pointmap regression for unconstrained image collections without requiring calibrated poses. VGGT \citep{wang2025vggt} generalizes this paradigm to variable numbers of views and jointly infers cameras, depth, point maps, and long-range tracks. Set-structured designs further improve scalability and robustness to view ordering, as in permutation-equivariant geometry learning \citep{wang2025pi}, and metric-oriented pipelines target consistent scale recovery across diverse captures \citep{keetha2025mapanything}. Complementary to multi-view geometry, large depth foundation models provide strong priors that can be distilled into reconstruction pipelines or used for initialization and regularization under limited supervision \citep{lin2025depth}.

Beyond static reconstruction, recent work pushes feedforward geometry toward large-scale and streaming settings, including reconstruction of large view sets with efficient global alignment \citep{xie2025fast3r}, persistent-state models for continuous 3D perception over long sequences \citep{wang2025cut3r,chen2025ttt3r}, and online spatial-memory reconstruction that incrementally integrates new views \citep{barisic2024spann3r}. In dynamic scenes, models built on DUSt3R-style representations predict time-varying structure with explicit motion awareness \citep{zhang2024monst3r} or training-free motion disentanglement \citep{chen2025easi3r}, enabling efficient reconstruction and novel view synthesis under substantial non-rigidity \citep{li2025wild3a}. While most foundation models are trained on predominantly clean imagery, early efforts have started to address adverse capture conditions by designing generalizable radiance-field pipelines for real-world degradations \citep{zhou2023nerflix,gupta2024gaura,wu2024rafe,yang2024drantal} and by proposing degradation-robust feedforward models \citep{liu2025lumos3d,wen2025splatbright}, suggesting an emerging direction toward feedforward reconstruction that remains reliable under noise, blur, and adverse conditions.

\subsection{Existing Degradation Benchmark}

Robust 3D reconstruction and novel view synthesis under real-world degradations depends critically on suitable datasets, yet most existing benchmarks target a single corruption type and only partially reflect the corruption patterns encountered in casual in-the-wild capture. Blur-oriented datasets \citep{ma2022deblur,lee2023exblurf,sun2024dyblurf,choi2025exploiting,lu2025bard,bui2025moblurf} commonly synthesize motion or defocus blur from high-frame-rate sharp videos, or obtain paired blurry and sharp observations via dual-camera rigs or repeated trajectories, which can simplify the underlying blur formation and limit diversity. BlurRF \citep{choi2025exploiting} provides rich multi-view coverage across motion and defocus blur, but a large portion of its samples are synthetic, leaving a gap for evaluating reconstruction robustness under real capture artifacts.

Datasets designed around transient occlusion and clutter \citep{snavely2006photo,sabour2023robustnerf,ren2024nerf} emphasize Internet photo collections or controlled tabletop scenes, but typically do not provide explicit low-quality and reference pairs, making faithful restoration-aware evaluation difficult. For illumination, RawNeRF \citep{mildenhall2022nerf} captures multi-view scenes in extreme darkness in RAW space, but does not include separate noise-free ground truth, while sRGB-based low-light benchmarks \citep{wang2023lighting,cui2024aleth,qu2024lush} offer paired low-light and normal-light views but remain limited in scene count and resolution, and often reflect simplified lighting setups compared to real scenes with complex, mixed illumination. Participating media datasets are scarcer and costly to acquire. Existing underwater \citep{levy2023seathru,muhammad2023underwater,wildflow2025sweet} and hazy-scene \citep{zhang2021learning,ramazzina2023scatternerf,jin2024reliable} datasets usually cover only a few scenes or provide restricted viewing angles. REVIDE \citep{zhang2021learning} delivers 48 aligned low-quality and reference video captures using a robot arm in 4 distinct scenes, but viewing angles remain limited by the robot's mechanical workspace.

RealX3D addresses these limitations by providing rich real captures with pixel-aligned reference views across diverse real-world degradations, combining high-resolution imagery with additional accessories, including geometric measurements, dedicated test views for NVS, and RAW sensor data to support both restoration and reconstruction benchmarking. A comprehensive comparison of existing degraded 3D datasets is shown in \Cref{tab:exist_dataset}.

\begin{figure*}[!tp]
\centering
\includegraphics[width=\textwidth]{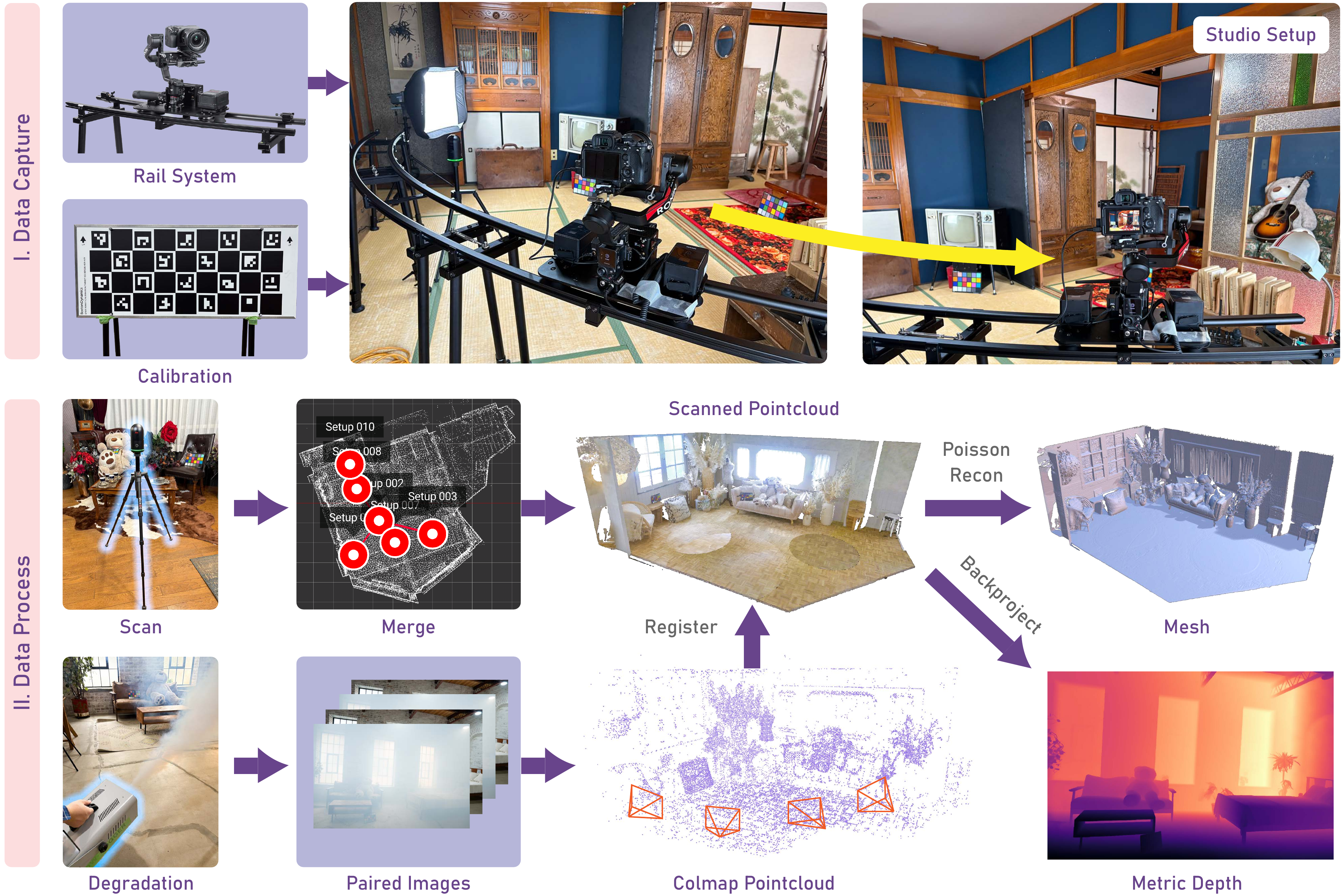}
\caption{Overview of our data acquisition and processing pipeline: we calibrate cameras; set up a rail-dolly and studio lighting to capture pixel-aligned LQ/GT pairs; scan the scene and register camera poses to world coordinates; then back-project to recover per-view metric depth, and reconstruct a high-quality mesh from the scans.}
\label{fig:data_collect}
\end{figure*}

\section{Unified Degradation Model}
To facilitate the acquisition of diverse real-world corruptions, we establish a unified degradation model. Specifically, we regard all degradations in RealX3D as perturbations of an underlying clean radiance map 
$J(x)$ captured under clear conditions. For a given degradation family $d$, the observed image $I_d$ can be written in a unified form:
\begin{equation}
    I_d(x) = \mathcal{B}_d \left[ T_d(x)J(x) + A_d(x)\right] + n_d(x)
\end{equation}
where $T_d(x) \in [0, 1]$ denotes the effective transmission of direct radiance, $A_d(x)$ collects parasitic radiance such as path radiance in participating media or extra view-dependent reflections, $\mathcal{B}_d$ is a non-trivial operator, and $n_d(x)$ subsumes sensor noise and residual nonlinearities. Illumination degradations correspond to a spatial-variant transmission factor $T_{\rm illu}$ that scales $J$ under different exposure settings. Scattering follows depth-dependent transmission $T_{\rm scat}$ and an in-scattered term $A_{\rm scat}(x)$. Occlusion and glass reflections are modeled via an effective visibility-modulated transmission $T_{\rm occ}(x)$ and additive occluder or reflective layers inside $A_{\rm occ}(x)$. Motion and defocus blur are captured by the blurring operator $\mathcal{B}_d$. 

Our unified model consolidates common real-world degradations into a single formulation, not only streamlining the data acquisition protocol, but also offering a principled foundation toward all-in-one in-the-wild 3D reconstruction.

\section{Data Acquisition Protocol and Processing}
In this section, we describe our data acquisition protocol and subsequent data processing pipelines. Specifically, we develop an acquisition system as illustrated in \Cref{fig:data_collect}, comprising a rail-based camera dolly, physical-degradation apparatus, and a high-end laser scanner. The system employs a high-precision programmable cart running on curved rails, delivering constant-velocity motion and repeatable viewpoints. A DJI RS4 gimbal stabilizes the camera to suppress micro-vibrations. Rails are mounted around 1 m, giving a lens height of 1.2–1.5 m and adequate parallax for indoor scenes. We use a Sony A74 with a 24–70 mm f/2.8 GM zoom, with focal length calibrated and fixed per scene. Low-quality views are produced via real physical degradations. For geometry, each scene is scanned with a high-end laser scanner. All data are acquired in professional studios with more than two 200W LED lights to maintain uniform illumination.

\subsection{GT and LQ Images}
\label{sec:lq_images}
We acquire sequences directly as individual RAW images rather than video frames. Using the rail-based camera dolly at a very low, constant speed, we trigger the shutter every second, typically obtaining over 400 images along a single trajectory. In scenes where the dolly cannot be deployed, we mount the camera on a fixed tripod and capture paired GT/LQ images with matched framing and exposure.

\subsubsection{Illumination Degradation}
We design two common illumination-related degradations: (i) consistent low light and (ii) low light with varying exposure. To ensure that GT images are not affected by noise or blurring in dark conditions, GT is always captured in a well-lit environment with a shutter speed of 1/10 s, and low-light LQ images are obtained by reducing exposure relative to this setting. For the consistent low-light condition, we fix the shutter speed at 1/400 s across all views to achieve extremely dark images. For varying low-light scenarios, inter-view brightness differences are introduced by capturing the same viewpoints at shutter speeds of 1/60, 1/160, 1/250, and 1/400 s, spanning roughly 0 to +2.7 EV. Beyond these physically captured low-light images, additional exposure settings can easily be synthesized from the RAW data.

\subsubsection{Scattering Degradation}
Existing simulation-based smoke or haze datasets commonly adopt the single-scattering Atmospheric Scattering Model (ASM) \citep{narasimhan2002vision}:
\begin{equation}
    I = J\cdot\text{exp}(-\beta z) + B^{\infty}\cdot(1-\text{exp}(-\beta z))
\end{equation}
Providing a clean image as $J$, the smoky or hazy image $I$ is synthesized by estimating per-pixel depth $z_i$ and applying a predefined scattering coefficient $\beta$ that counts in-scatter ambient light $B^{\infty}$ to generate the degraded views. However, this approach assumes only single scattering along the LoS and ignores the attenuation of light before it reaches the surface. In real-world scattering scenarios, the apparent object radiance $J$ is substantially lower than that of the ideal scattering-free $\hat{J}$, because the incident light is attenuated as it propagates through the medium from the light source to the scene surfaces. Moreover, scattering often occurs multiple times. As a result, such synthetic datasets exhibit a large domain gap from real-world situations.

To collect real smoke data, we leverage our precise camera dolly system. Capture is performed in sealed indoor scenes: we first record GT multiview images along the rail, then generate persistent smoke using a 1200 W smoke machine that atomizes liquid into dense aerosol particles. After the smoke diffuses uniformly, we recapture the same trajectory to obtain scattering-degraded LQ images.

\begin{figure*}[!tp]
\centering
\includegraphics[width=1\textwidth]{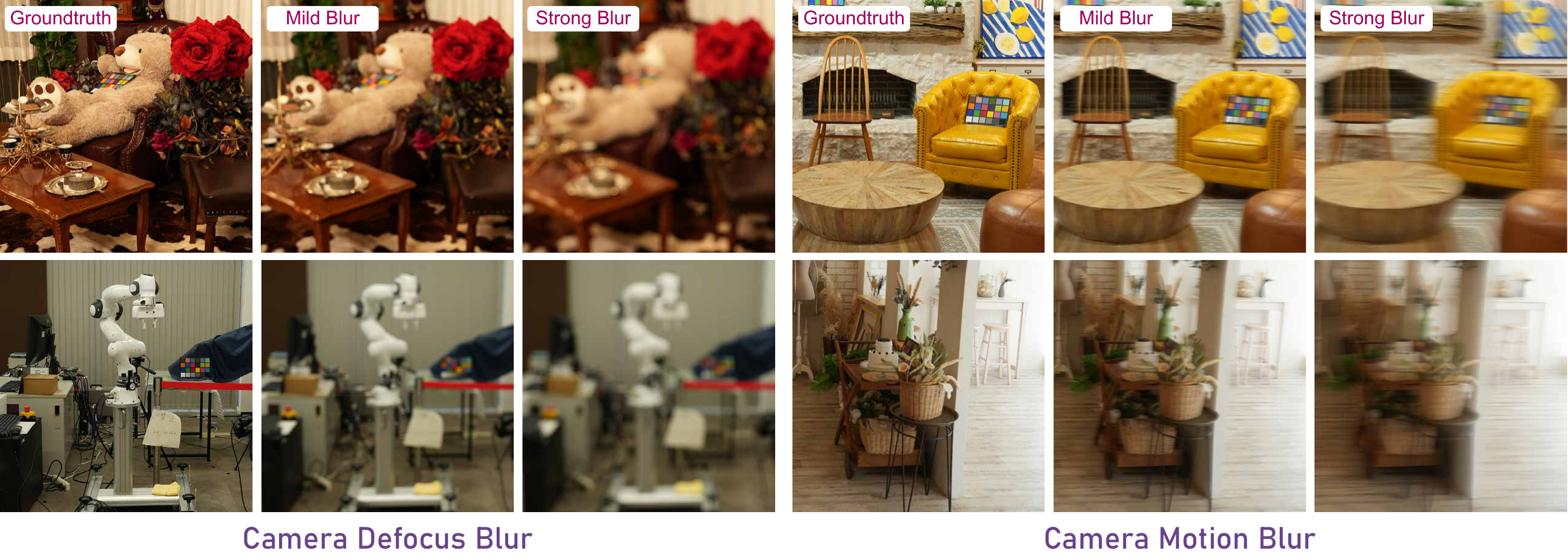}
\caption{Visualization of blur severity at two levels. Left: camera defocus blur under mild and strong settings, with focal distances of 0.6 meters and 0.4 meters. Right: camera motion blur under mild and strong settings, with 2 and 5 centimeters of camera displacement during exposure, respectively.}
\label{fig:blur_example}
\end{figure*}

\subsubsection{Occlusion Degradation}
We introduce two types of occlusion: (i) transient occluders in the scene, and (ii) reflection-induced artifacts. For transient occluders, we adopt two acquisition settings that are randomly applied during data collection. One setting uses static objects that stay fixed during each exposure but are rearranged between viewpoints. The other introduces fast-moving objects that create motion-blurred streaks and ghosting. Static occluders can be separated using a pretrained segmentation network to obtain dynamic masks, whereas motion-blurred occluders lack clear boundaries and are therefore difficult to segment reliably. Reflections provide another form of occlusion-related degradation: by mounting a transparent glass plate with 92\% transmittance in front of the lens, we create additional reflective layers whose radiance introduces view-dependent, inconsistent artifacts in each image, which are even harder for detection-based methods to recognize. Examples of the dynamic and reflective occlusion are visualized in \Cref{fig:teaser}.

\subsubsection{Blurring Degradation}
We consider two common blur degradations: (i) defocus blur and (ii) camera motion blur. For defocus blur, existing datasets typically shift focus to either the foreground or the background, leaving one region sharp while the other is partially blurred. In contrast, we introduce a global out-of-focus variant. In our captures, scenes are normally focused at 3–5 m; for the defocus setting, we deliberately misfocus the lens to 0.6 m (mild) and 0.4 m (strong), and acquire both levels consistently for every scene.

For camera motion blur, our goal is to obtain pixel-aligned GT/LQ pairs while remaining faithful to the physical image-formation process. Motion blur arises when the camera moves during exposure and the sensor integrates radiance along the motion path. We first reconstruct a clean 3D scene from the sharp GT images and estimate a calibrated camera trajectory along the dolly path. For each target frame, we assume constant-speed motion and define a blur path length that determines the camera travel during the exposure. Specifically, we synthesize two blur levels by integrating over path lengths of 2 cm (mild) and 6 cm (strong) preceding the target pose. Along each path segment, we uniformly sample 64 intermediate poses, render the corresponding views, and integrate them with the target GT image under the standard exposure-integration model for a moving camera. These controlled path lengths yield two physically consistent motion-blur strengths. Examples of camera defocus and motion blur are shown in \Cref{fig:blur_example}.

\begin{figure*}[!tp]
\centering
\includegraphics[width=1\textwidth]{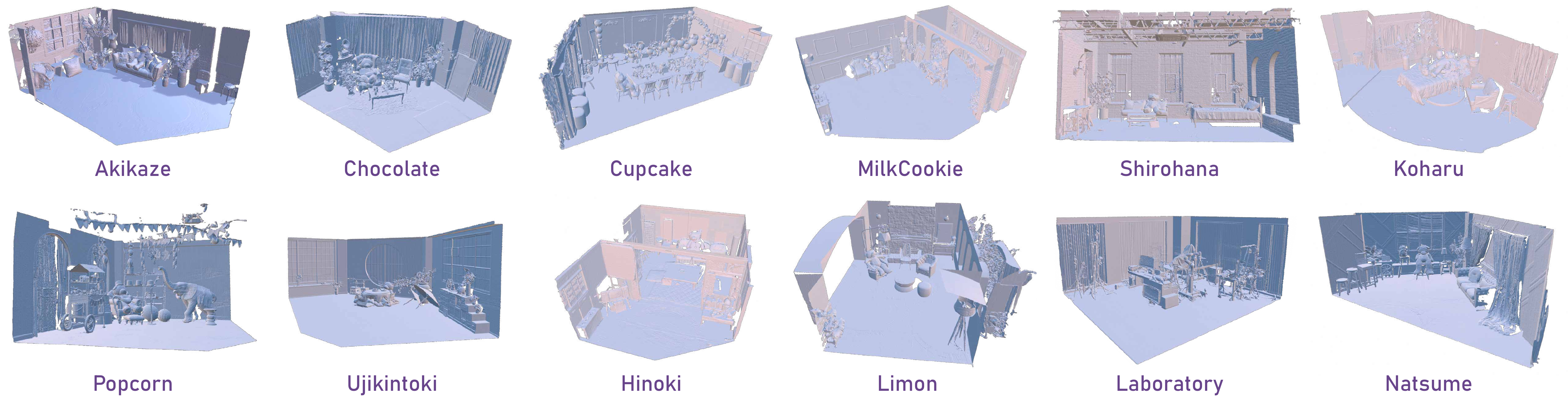}
\caption{Visualization of representative RealX3D scene meshes reconstructed from dense laser-scanned point clouds.}
\label{fig:mesh}
\end{figure*}

\subsection{Pose Estimation}
\label{sec:calibration_pose}

To enhance dataset diversity and accommodate focal adjustments across scenes, we use a zoom lens with a predefined range of 24–34 mm. Before capturing images, camera intrinsics are calibrated for each focal length using a 4×9 ChArUco board. During each scene capture, the focal length remains fixed. Because degradations prevent accurate pose recovery from LQ images using COLMAP \citep{schonberger2016structure}, we leverage pixel-aligned LQ/GT pairs and estimate camera poses on the corresponding GT images with the calibrated intrinsics, followed by undistortion of both LQ and GT into the pinhole camera model. As detailed in \Cref{sec:pointcloud_registration}, the COLMAP-derived poses and feature-based point cloud are registered to the laser-scan data, yielding a final root-mean-square error of 1.2 cm in complex indoor environments, demonstrating the accuracy of the estimated poses.

\subsection{Laser Scans For World-scale Geometry Measure}
\label{sec:scan}

The BLK360 G2 high-end scanner offers a native precision of 4 mm at 10 m and captures about 50 million points per scan. In complex indoor environments, semi-transparent and reflective surfaces can introduce noise, so for each individual scan, we remove points with reflectance intensity below 24 to suppress unreliable measurements. Each scene is scanned at least five times in HDR mode, projecting color information onto the point cloud. All scans are then registered and fused, followed by 5mm-uniform subsampling to obtain a dense cloud. Finally, we apply Poisson surface reconstruction to generate a mesh and decimate it by 20 percent to remove redundant vertices, as shown in \Cref{fig:mesh}.

\subsection{Point Cloud Registration}
\label{sec:pointcloud_registration}

We register the sparse COLMAP point cloud with the dense laser-scan point cloud. To handle scale mismatch, we first manually select 5–8 correspondences to perform a coarse alignment, then refine with ICP \citep{besl1992method} to obtain an accurate rigid transform from the COLMAP coordinates to the scan’s world coordinates. Applying this transform to the COLMAP camera poses places them in real-world coordinates. Using the transformed poses, we render the reconstructed mesh to obtain metric depth for each view. The world-aligned poses and dense point cloud provide accurate labels for geometric evaluation.

\subsection{Data Source}
Using our acquisition protocol and processing pipeline, RealX3D provides 2,407 paired low-quality and reference images, along with the same number of corresponding RAW captures. The dataset is collected across 15 indoor rooms and organized into 55 distinct scenes spanning seven degradation types. The current release includes defocus or camera motion blur with 8 scenes and 271 pairs, where each scene is captured at two blur severity levels; dynamic occlusion with 8 scenes and 271 pairs; reflection with 8 scenes and 271 pairs; extreme low light with 9 scenes and 319 pairs; low-light exposure variation with 9 scenes and 319 pairs; and smoke scattering with 5 scenes and 143 pairs.

We further provide laser-scanned point clouds with 5 mm point spacing, along with calibrated camera intrinsics and extrinsics. Each view is paired with a metric depth map stored as a 16-bit PNG in millimeters, and each scene includes a colored mesh reconstructed from the scanned point clouds.

\section{Performance Metrics}
We design our evaluation to match the fundamentally different pipelines of optimization-based dense reconstruction methods and feedforward foundation models. Optimization-based methods typically depend on SfM pipelines \citep{schonberger2016structure} to estimate camera poses and a sparse point cloud, followed by dense reconstruction and rendering. Under real-world degradations, especially the severe pixel corruption in RealX3D, conventional SfM often fails, which prevents a fair assessment of the dense reconstruction stage. To decouple SfM failure from the evaluation, we fix camera poses to the ground-truth poses obtained from the pixel-aligned GT views, use the LQ images as input, and measure photometric fidelity of each method to reconstruct and restore scene appearance on both training views and NVS.

In contrast, feedforward foundation models take LQ images as input, and simultaneously predict camera poses and 3D geometry. We therefore evaluate robustness under degradations using pose accuracy and geometry quality metrics, which directly reflect how well the model can infer reliable structure and viewpoint parameters from corrupted observations.

\begin{table*}[!tp]
\centering
\caption{Quantitative comparisons of average training-view and NVS performance across all scenes for each real-world degradation setting. For defocus and motion blur, results are reported under the strong-blur setting. Detailed per-scene performance on low-light, varying exposure, smoke, dynamic occlusion, reflection, camera motion blur and camera defocs blur is outlined in \Cref{tab:sup_lowlight}, \Cref{tab:sup_exposure}, \Cref{tab:sup_scattering}, \Cref{tab:sup_dynamic}, \Cref{tab:sup_reflection}, \Cref{tab:sup_motion100}, \Cref{tab:sup_motion300}, \Cref{tab:sup_defocus60}, and \Cref{tab:sup_defocus40}, respectively.}
\setlength{\tabcolsep}{0pt}
\BigColBegin
\noalign{\vskip 1pt}%
\TypeFirst{Low-light}{5}{3DGS \citep{kerbl20233d}}{6.58}{6.66}{0.060}{0.058}{0.656}{{0.659}}
\TypeNext{Aleth-NeRF \citep{cui2024aleth}}{{12.98}}{{12.99}}{0.450}{{0.445}}{0.706}{0.704}
\TypeNext{Luminance-GS \citep{cui2025luminance}}{10.89}{10.05}{{0.531}}{0.433}{{0.640}}{0.708}
\TypeNext{LITA-GS \citep{zhou2025lita}}{\textbf{15.63}}{\textbf{15.57}}{{0.542}}{{0.542}}{\textbf{0.483}}{\textbf{0.488}}
\TypeNext{I2-NeRF \citep{liu2025i2nerf}}{{15.55}}{{15.51}}{\textbf{0.584}}{\textbf{0.568}}{{0.514}}{{0.532}}
\TypeSep
\TypeFirst{VaryExp}{3}{3DGS \citep{kerbl20233d}}{{7.29}}{{7.39}}{{0.124}}{{0.131}}{{0.623}}{{0.643}}
\TypeNext{Luminance-GS \citep{cui2025luminance}}{{13.22}}{{14.51}}{{0.451}}{\textbf{0.568}}{{0.633}}{{0.556}}
\TypeNext{LITA-GS \citep{zhou2025lita}}{\textbf{16.06}}{\textbf{15.83}}{\textbf{0.563}}{{0.546}}{\textbf{0.467}}{\textbf{0.485}}
\TypeSep
\TypeFirst{Smoke}{5}{3DGS \citep{kerbl20233d}}{{9.87}}{{9.76}}{\textbf{0.517}}{\textbf{0.499}}{\textbf{0.629}}{\textbf{0.659}}
\TypeNext{SeaThruNeRF \citep{levy2023seathru}}{7.58}{7.55}{{0.467}}{{0.464}}{{0.679}}{{0.683}}
\TypeNext{Watersplatting \citep{li2025watersplatting}}{\textbf{10.74}}{\textbf{10.78}}{0.445}{0.445}{0.720}{0.723}
\TypeNext{SeaSplat \citep{yang2025seasplat}}{{10.46}}{{10.42}}{{0.452}}{{0.446}}{0.768}{0.774}
\TypeNext{I2-NeRF \citep{liu2025i2nerf}}{8.39}{8.40}{0.283}{0.283}{{0.696}}{{0.699}}
\TypeSep
\TypeFirst{Dynamic}{5}{3DGS \citep{kerbl20233d}}{{22.29}}{19.83}{{0.829}}{0.739}{{0.259}}{{0.342}}
\TypeNext{GS-W \citep{zhang2024gaussian}}{20.51}{{21.71}}{0.760}{{0.741}}{0.369}{0.391}
\TypeNext{Wild3A \citep{li2025wild3a}}{19.91}{15.67}{0.719}{0.541}{0.398}{0.552}
\TypeNext{SpotLessSplat \citep{sabour2025spotlesssplats}}{\textbf{28.68}}{\textbf{26.28}}{\textbf{0.864}}{\textbf{0.841}}{{0.258}}{{0.274}}
\TypeNext{DeSplat \citep{wang2025desplat}}{{25.64}}{{23.34}}{{0.855}}{{0.810}}{\textbf{0.212}}{\textbf{0.248}}
\TypeSep
\TypeFirst{Reflection}{5}{3DGS \citep{kerbl20233d}}{{24.07}}{21.60}{{0.841}}{{0.776}}{{0.220}}{{0.284}}
\TypeNext{GS-W \citep{zhang2024gaussian}}{22.51}{{23.06}}{0.783}{0.757}{0.324}{0.360}
\TypeNext{Wild3A \citep{li2025wild3a}}{23.42}{18.59}{0.774}{0.620}{0.295}{0.440}
\TypeNext{SpotLessSplat \citep{sabour2025spotlesssplats}}{\textbf{26.06}}{\textbf{24.52}}{{0.843}}{\textbf{0.825}}{{0.271}}{{0.286}}
\TypeNext{DeSplat \citep{wang2025desplat}}{{25.02}}{{23.21}}{\textbf{0.845}}{{0.804}}{\textbf{0.211}}{\textbf{0.246}}
\TypeSep
\TypeFirst{Motion Blur}{6}{3DGS \citep{kerbl20233d}}{{20.33}}{{19.41}}{{0.663}}{{0.637}}{{0.484}}{0.508}
\TypeNext{DeBlurring-3DGS \citep{zhang2024gaussian}}{\textbf{20.64}}{\textbf{20.32}}{\textbf{0.702}}{\textbf{0.690}}{\textbf{0.448}}{\textbf{0.455}}
\TypeNext{Deblur-GS \citep{chen2024deblur}}{18.04}{17.85}{0.554}{0.553}{0.548}{0.548}
\TypeNext{Bad-Gaussians \citep{zhao2024bad}}{19.28}{18.85}{0.590}{0.587}{0.514}{0.520}
\TypeNext{BAGS \citep{peng2024bags}}{18.75}{18.39}{0.575}{0.568}{0.495}{{0.502}}
\TypeNext{CoCoGaussian \citep{lee2025cocogaussian}}{{20.14}}{{19.90}}{{0.638}}{{0.635}}{{0.455}}{{0.461}}
\TypeSep
\TypeFirst{Defocus Blur}{6}{3DGS \citep{kerbl20233d}}{{20.83}}{{19.79}}{\textbf{0.631}}{\textbf{0.616}}{{0.582}}{0.599}
\TypeNext{DeBlurring-3DGS \citep{zhang2024gaussian}}{19.82}{19.47}{0.608}{0.601}{0.596}{0.602}
\TypeNext{Deblur-GS \citep{chen2024deblur}}{18.13}{17.97}{0.584}{0.580}{0.604}{0.608}
\TypeNext{Bad-Gaussians \citep{zhao2024bad}}{19.35}{18.86}{0.606}{{0.602}}{0.588}{{0.596}}
\TypeNext{BAGS \citep{peng2024bags}}{\textbf{21.01}}{\textbf{20.58}}{{0.610}}{0.593}{\textbf{0.555}}{\textbf{0.563}}
\TypeNext{CoCoGaussian \citep{lee2025cocogaussian}}{{20.40}}{{20.07}}{{0.623}}{{0.615}}{{0.572}}{{0.580}}
\BigColEnd
\label{tab:photometric_eval}
\end{table*}

\subsection{Photometric Fidelity}
We evaluate appearance restoration and view-consistent rendering using PSNR \citep{hore2010image}, SSIM \citep{wang2004image}, and LPIPS \citep{zhang2018unreasonable}. For each method, we render the corresponding view and compute metrics against the paired, pixel-aligned reference images for both training views and held-out novel testing views. PSNR measures per-pixel reconstruction error in logarithmic scale and is sensitive to absolute intensity differences. SSIM compares local luminance, contrast, and structure to reflect structural consistency. LPIPS evaluates perceptual distance in a deep feature space and is more sensitive to texture and semantic discrepancies that may not be captured by pixel-wise metrics. Higher PSNR and SSIM, and lower LPIPS indicate better photometric fidelity. We report results averaged over views for each scene.

\subsection{Pose Accuracy}
We evaluate camera pose accuracy using the area under curve (AUC) of pose errors under multiple thresholds, reported as AUC@5, AUC@10, and AUC@20. For each estimated pose, we compute its error with respect to the GT pose, typically combining rotation and translation discrepancies into a single scalar pose error. We then compute the cumulative accuracy curve by measuring the fraction of poses whose error is below a threshold, and calculate the area under this curve up to the specified cutoff. Higher AUC indicates more accurate and stable pose estimation under degradations.

\subsection{Geometry Accuracy}
We evaluate geometric quality from two complementary perspectives, capturing both per-view depth consistency and scene-level surface accuracy.

At the view level, we compute the Depth L1 error between the predicted depth map and the GT metric depth for each view. We adopt a standard monocular depth evaluation practice to resolve scale ambiguity. Specifically, we compute the median depth over valid pixels for both predicted and GT depth maps, rescale the predicted depth by the median-ratio to match the real-world scale, and then compute the L1 error. We average the depth errors over views to obtain per-scene depth accuracy. Lower Depth L1 error represents higher depth prediction accuracy. 

At the scene level, we compare the predicted point cloud to the GT mesh reconstructed from dense laser scans. We first align the predicted point cloud to the real-world coordinate system using ICP \citep{besl1992method}. After alignment, we compute point-to-surface distances with a maximum error threshold of 5cm to quantify both geometric accuracy and surface coverage. Accuracy measures the distance from predicted points to the GT surface, while completeness measures the distance from GT surface samples to the predicted point set. We additionally report the F1 score, defined as the harmonic mean of precision and recall under the same threshold of 5cm, where precision is the fraction of predicted points that lie within the error threshold, and recall is the fraction of GT surface samples that lie within the error threshold of the prediction. Higher F1 scores indicate better overall geometric reconstruction quality

\section{Experiment}

\subsection{Photometric Evaluation}
We conduct a comprehensive task-specific evaluation on RealX3D and report averaged quantitative results for each degradation type in \Cref{tab:photometric_eval}. In each setting, we also report the performance of vanilla 3DGS \citep{kerbl20233d} as a reference.

\begin{figure*}[!tp]
\centering
\includegraphics[width=\textwidth]{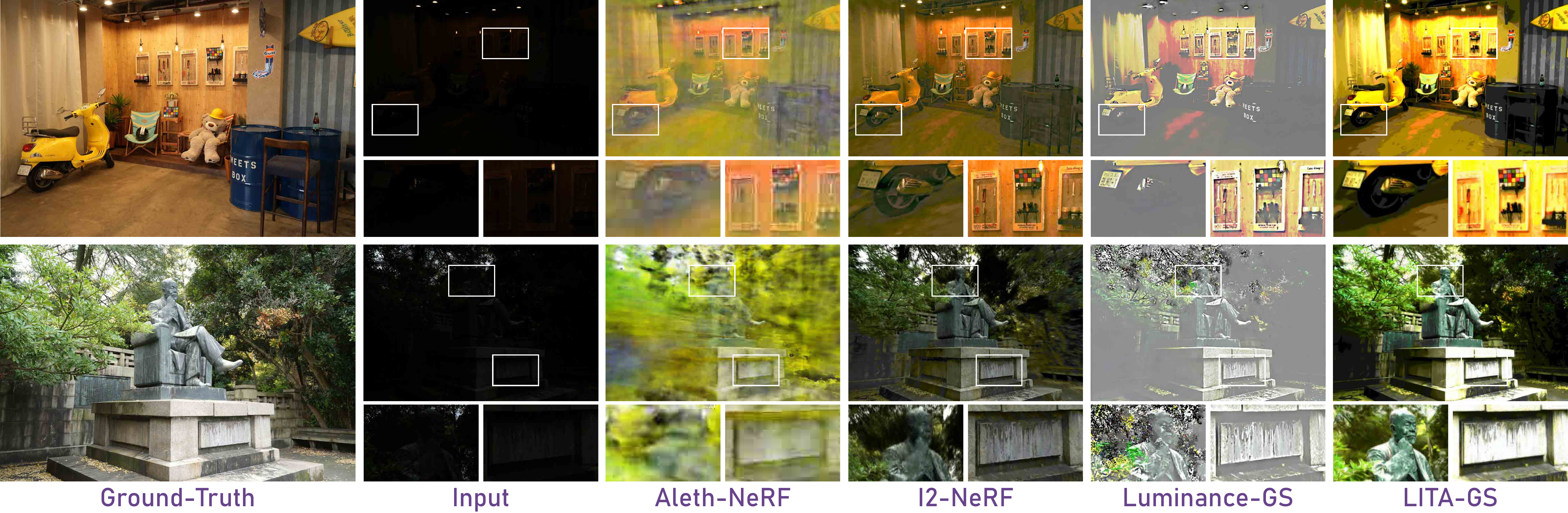}
\caption{Qualitative comparison of baseline methods on selected \textbf{\textit{low-light}} scenes in the RealX3D benchmark.}
\label{fig:render_lowlight}
\end{figure*}

\begin{figure*}[!tp]
\centering
\includegraphics[width=\textwidth]{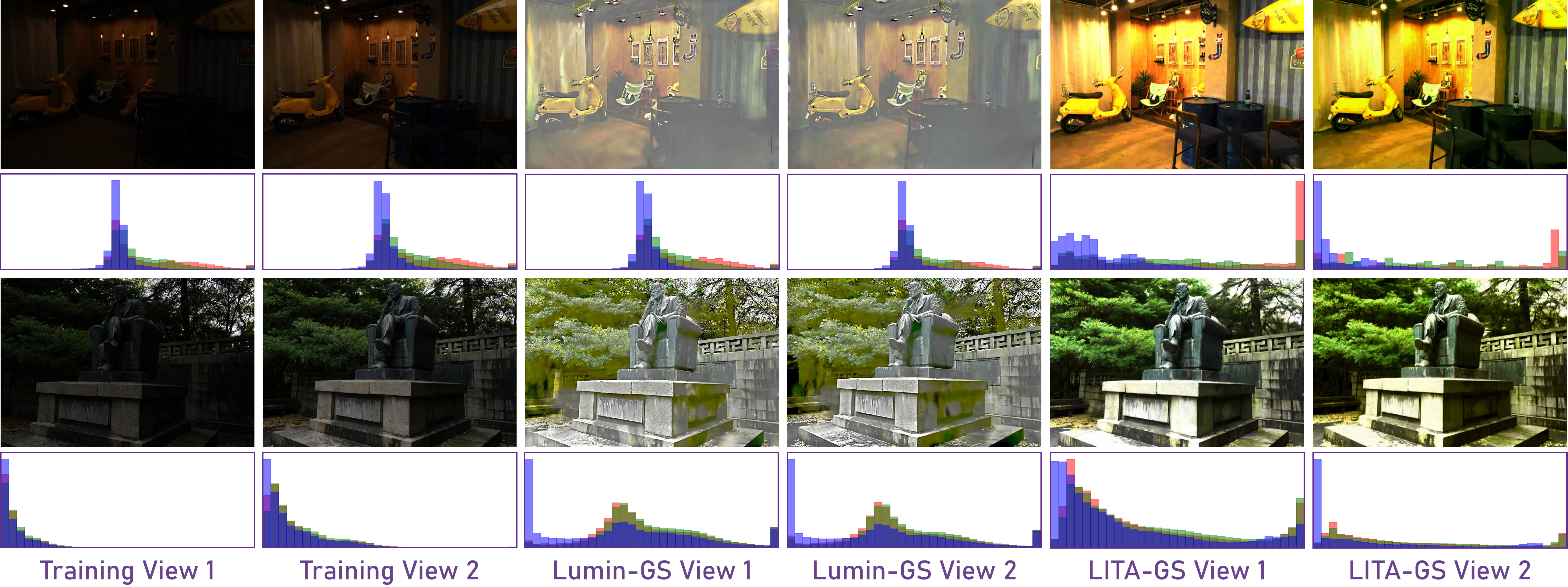}
\caption{Qualitative comparison of baseline methods on selected \textbf{\textit{varying exposure}} scenes in the RealX3D benchmark. We show two adjacent example training views and the corresponding rendered restoration results for each method. The second row visualizes the per-image pixel intensity histograms of the rendered outputs.}
\label{fig:render_exposure}
\end{figure*}

\subsubsection{Extreme Low-light Restoration}
We evaluate representative sRGB-based low-light 3D reconstruction methods, including Aleth-NeRF \citep{cui2024aleth}, I2-NeRF \citep{liu2025i2nerf}, Luminance-GS \citep{cui2025luminance}, and LITA-GS \citep{zhou2025lita}. These methods span both NeRF-based and 3DGS-based formulations, and represent the current state of the art for low-light novel view synthesis.

As shown in \Cref{tab:photometric_eval}, all evaluated methods exhibit substantial performance degradation under extreme low-light conditions. Compared with the results reported on existing benchmarks such as LOM \citep{cui2024aleth} and LLNeRF \citep{wang2023lighting}, performance on RealX3D drops markedly in terms of perceptual metrics. This performance gap highlights the increased difficulty posed by RealX3D, which features complex scene geometry, denser multi-view capture, wider viewpoint diversity, and spatially non-uniform light setup. These factors jointly exacerbate the ambiguity of photometric cues under severe illumination degradation, making both radiance estimation and geometry reconstruction significantly more challenging.

Qualitative results in \Cref{fig:render_lowlight} further reveal characteristic failure patterns across methods. Aleth-NeRF \citep{cui2024aleth} and I2-NeRF \citep{liu2025i2nerf} tend to produce under-exposed renderings with suppressed details, particularly in shadowed regions, indicating insufficient recovery of global illumination. Luminance-GS \citep{cui2025luminance} partially improves brightness but often suffers from contrast collapse and spatially inconsistent luminance, leading to flat appearances and loss of fine structure. LITA-GS \citep{zhou2025lita} achieves the superior quantitative performance among the evaluated methods; however, its reconstructions frequently exhibit noticeable hue distortion and color shifts. These artifacts are likely caused by instability in tone-mapping applied post-optimization, which disrupts color fidelity. Across all methods, such photometric errors are not confined to appearance alone: they propagate from rendered views into the reconstructed scene geometry, resulting in inconsistent structures and visible floaters in both NeRF and 3DGS representations.

\begin{figure*}[!tp]
\centering
\includegraphics[width=\textwidth]{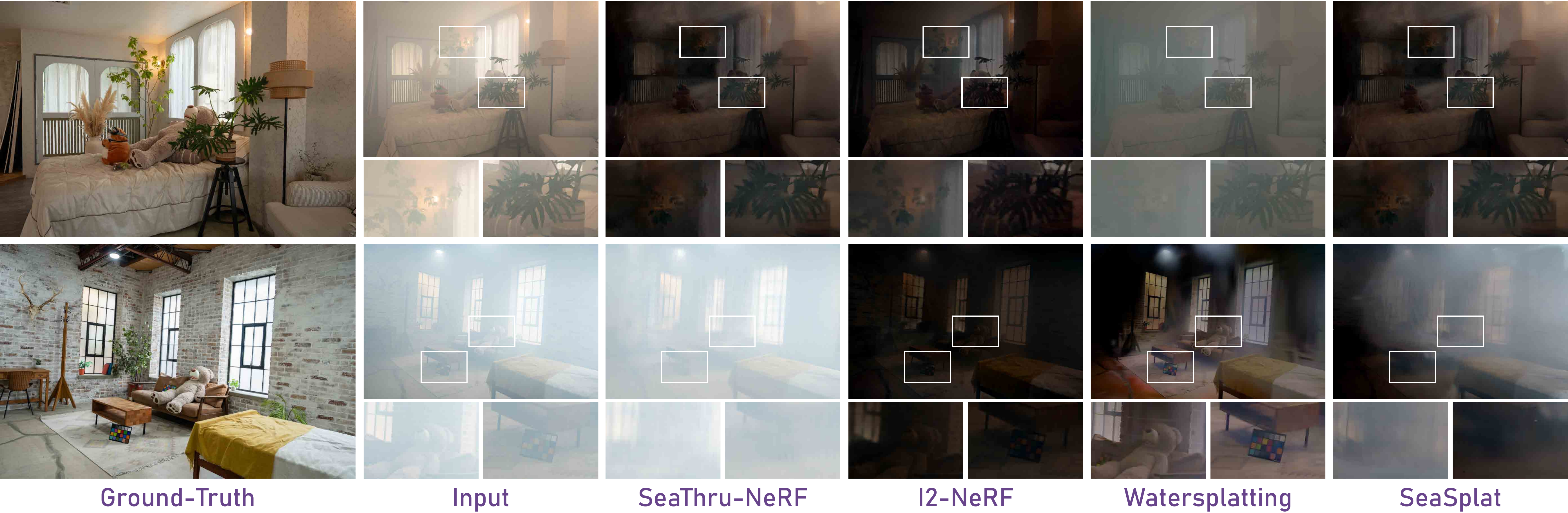}
\caption{Qualitative comparison of baseline methods on selected \textbf{\textit{smoke}} scenes in the RealX3D benchmark.}
\label{fig:render_smoke}
\end{figure*}

\subsubsection{Low-light with Varying Exposure}

Real-world image capture frequently exhibits illumination inconsistency across viewpoints due to temporal exposure variation, automatic camera control, and dynamic lighting conditions. Such effects are particularly pronounced in low-light scenarios, where exposure adjustments are aggressively applied to compensate for insufficient photon counts, resulting in substantial appearance shifts across views. These variations introduce significant challenges for multi-view reconstruction, as identical scene points may be observed under markedly different radiometric conditions.

In this setting, we explicitly model view-dependent exposure variation by assigning different exposure times across viewpoints within the same scene. Robust reconstruction methods are therefore required to disentangle intrinsic scene appearance from exposure-induced intensity changes by learning illumination-invariant representations \citep{niemeyer2025learning}. Unlike HDR reconstruction \citep{huang2022hdr,cai2024hdr,jin2024lighting} that aims to recover high dynamic range radiance from bracketed low dynamic range observations, our task focuses on evaluating the robustness of low-light 3D reconstruction and NVS under exposure inconsistency, without assuming access to exposure-aligned inputs or radiometrically calibrated supervision.

We evaluate recent illumination-aware methods, including Luminance-GS \citep{cui2025luminance} and LITA-GS \citep{zhou2025lita}, which are designed to mitigate view-dependent illumination shifts and improve cross-view appearance consistency. As reported in \Cref{tab:photometric_eval}, both methods exhibit notably low performance in the varying-exposure setting. Qualitative results in \Cref{fig:render_exposure} show the rendering outcomes. Luminance-GS \citep{cui2025luminance}produces view-consistent renderings in terms of global brightness, but suffers from incorrect enhancement and contrast collapse. In contrast, LITA-GS \citep{zhou2025lita} yields visually sharper results with superior brightness recovery; however, its per-view pixel histograms indicate substantial distribution shifts across viewpoints, revealing color and tone inconsistencies.

The evaluation results demonstrate that varying exposure under low-light conditions remains a challenging and underexplored setting for 3D reconstruction. Even methods explicitly designed for illumination robustness struggle to simultaneously maintain exposure invariance, color fidelity, and view-consistent appearance.

\begin{figure*}[!tp]
\centering
\includegraphics[width=\textwidth]{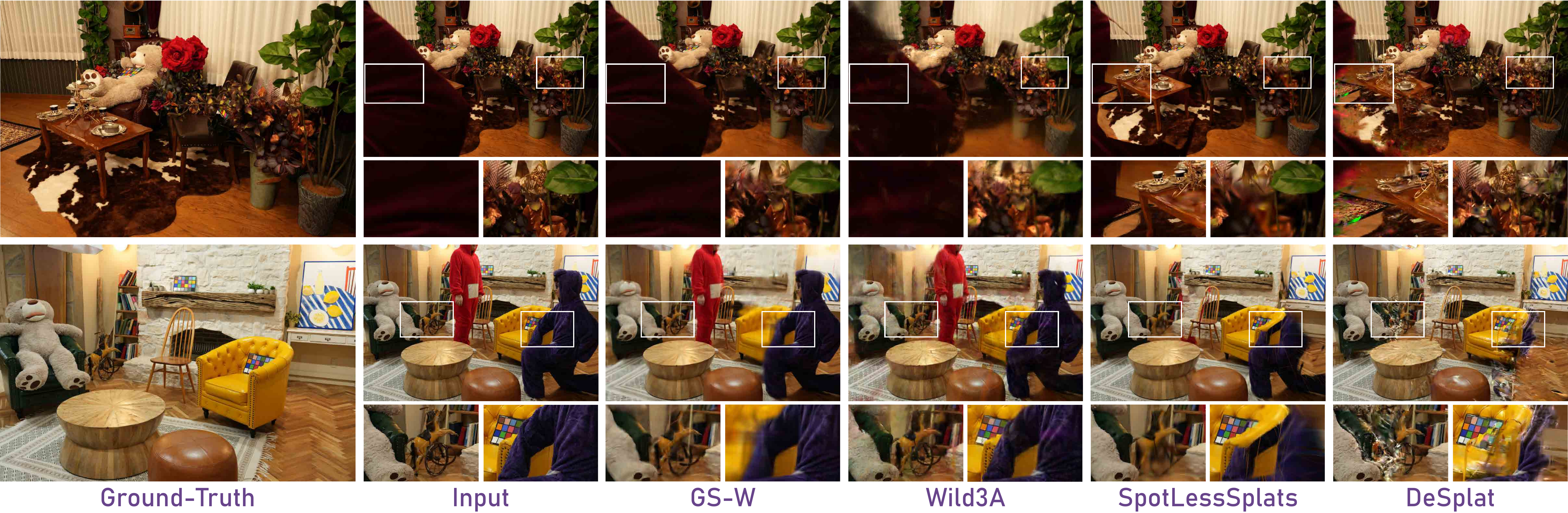}
\caption{Qualitative comparison of baseline methods on selected \textbf{\textit{dynamic occlusion}} scenes in the RealX3D benchmark.}
\label{fig:render_dynamic}
\end{figure*}

\begin{figure*}[!tp]
\centering
\includegraphics[width=\textwidth]{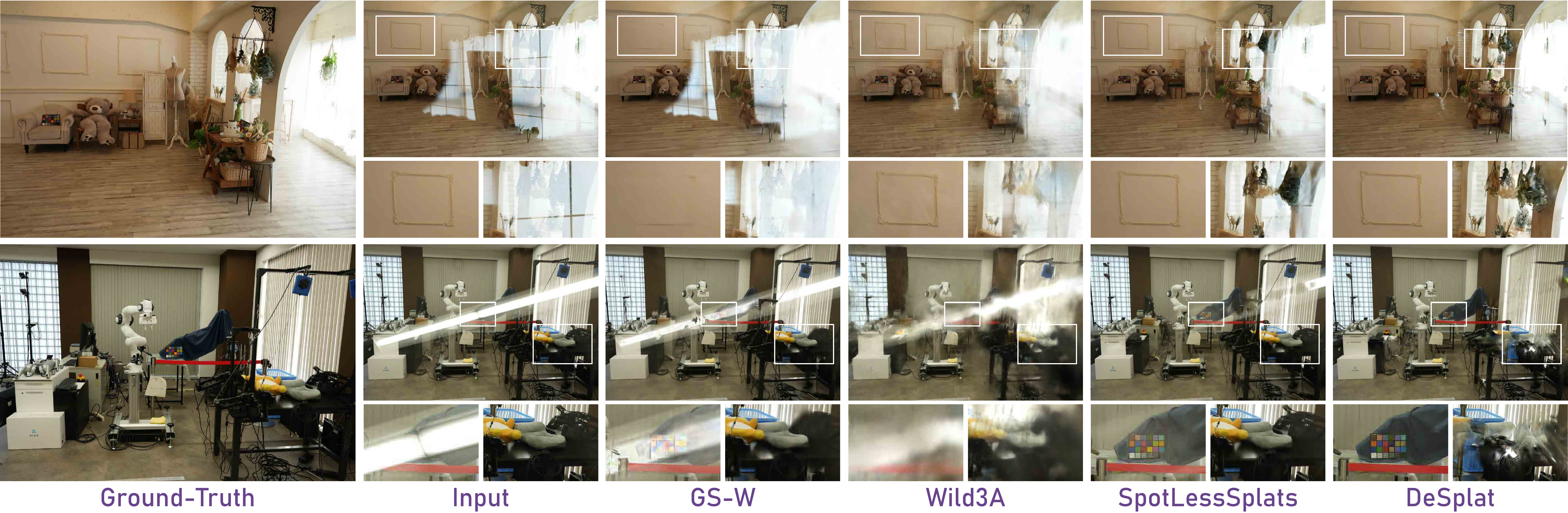}
\caption{Qualitative comparison of baseline methods on selected \textbf{\textit{reflection}} scenes in the RealX3D benchmark.}
\label{fig:render_reflection}
\end{figure*}

\subsubsection{Smoke Scattering}
Since atmospheric scattering and underwater scattering share similar physical mechanisms, both dominated by in-scattering and attenuation and differing primarily in wavelength dependency, we evaluate recent scattering-aware 3D reconstruction methods that explicitly model participating media. Specifically, we consider SeaThru-NeRF \citep{levy2023seathru}, I2-NeRF \citep{liu2025i2nerf}, Watersplatting \citep{li2025watersplatting}, and SeaSplat \citep{yang2025seasplat}, all of which incorporate in-scattered radiance into the rendering process. Although certain methods are primarily developed for underwater scenes, their formulations can transfer to smoke-like scattering conditions captured by the RealX3D benchmark.

Unlike the strong performance reported on previous synthetic benchmarks \citep{levy2023seathru,liu2025i2nerf,li2025watersplatting}, the quantitative results in \Cref{tab:photometric_eval} show that real-world scattering remains substantially more challenging, and all evaluated baselines exhibit remarkable dropbacks. As shown in \Cref{fig:render_smoke}, the existing methods often struggle to disentangle the scattering medium from the underlying scene. Some baselines fail to adequately account for in-scattered radiance, producing over-attenuated renderings with reduced brightness and missing details. Others mistakenly attribute scattering effects to scene appearance or geometry, suppressing true surface radiance and yielding washed-out and blurred reconstruction.

These failures are amplified by the characteristics of real smoke, including spatially varying density, non-uniform airlight, and strong view-dependent attenuation, which are typically simplified or absent in synthetic settings. Consequently, photometric errors propagate into geometry estimation, leading to degraded novel view synthesis. These results highlight a substantial domain gap between synthetic scattering benchmarks and real-world environments, underscoring the need for realistic datasets to drive progress in scattering-robust 3D reconstruction.

\begin{figure*}[!tp]
\centering
\includegraphics[width=\textwidth]{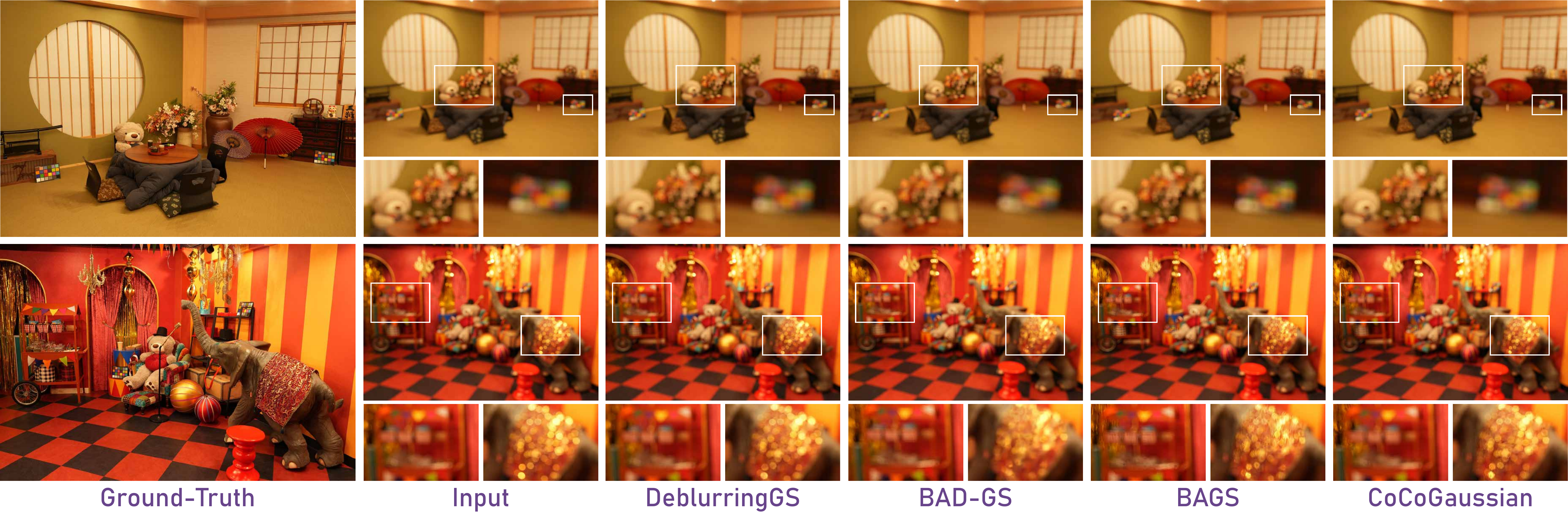}
\caption{Qualitative comparison of baseline methods on selected \textbf{\textit{defocus blur}} scenes in the RealX3D benchmark.}
\label{fig:render_defocus}
\end{figure*}

\begin{figure*}[!tp]
\centering
\includegraphics[width=\textwidth]{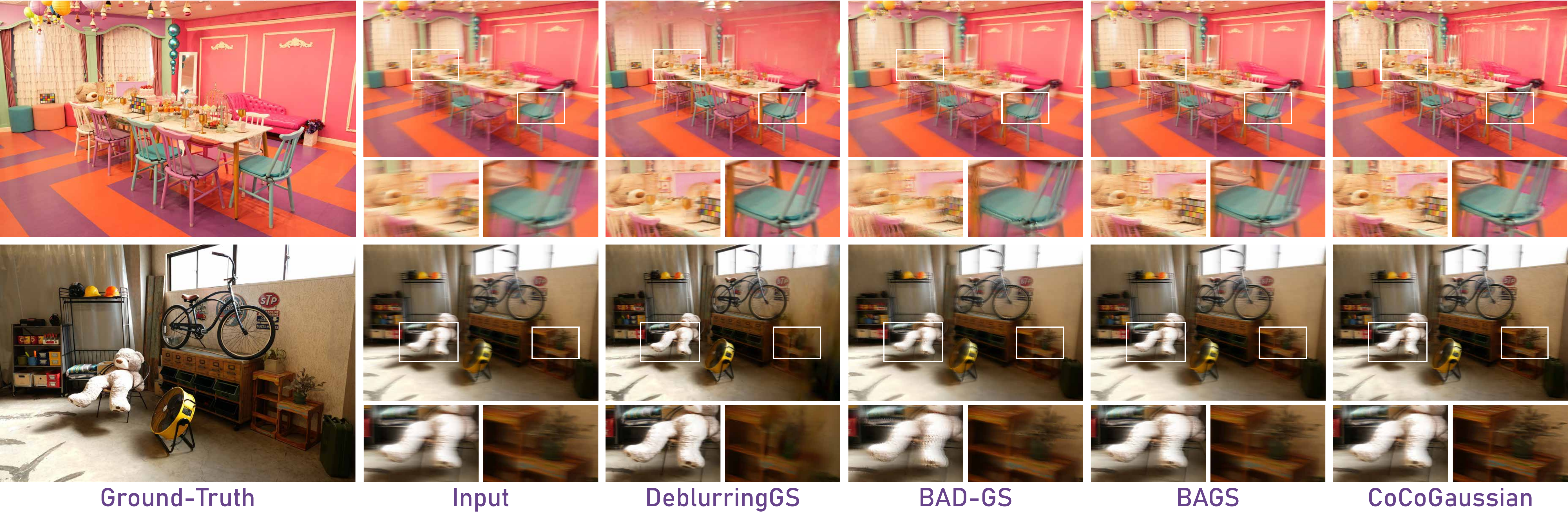}
\caption{Qualitative comparison of baseline methods on selected \textbf{\textit{motion blur}} scenes in the RealX3D benchmark.}
\label{fig:render_motion}
\end{figure*}

\begin{table*}[tb]
\centering
\small
\caption{Quantitative comparisons of averaged pose accuracy of feedforward models. Percentages denote the relative decrease with respect to the corresponding metric on clean views.}
\setlength{\tabcolsep}{0pt}
\setlength{\extrarowheight}{3pt}
\begin{tabular*}{\linewidth}{@{\extracolsep{\fill}}l @{\hspace{4pt}}|ccc ccc ccc}
\hline
\textbf{Methods} &
\multicolumn{3}{c}{AUC@5$\uparrow$} &
\multicolumn{3}{c}{AUC@10$\uparrow$} &
\multicolumn{3}{c}{AUC@20$\uparrow$} \\
\cmidrule(lr){2-4} \cmidrule(lr){5-7} \cmidrule(lr){8-10}
\noalign{\vskip -3pt}
 & Clean & Degrade & Error & Clean & Degrade & Error & Clean & Degrade & Error \\
\hline
VGGT \citep{wang2025vggt} & 86.13
& \textbf{82.71} & \textbf{4\%} & 92.98
& \textbf{91.32} & \textbf{2\%} & 96.49
& \textbf{95.66} & \textbf{1\%} \\
Pi3 \citep{wang2025pi} & 86.57
& 76.15 & 12\% & 93.28
& 87.92 &6\% & 96.64
& 93.95 & 3\% \\
MapAnything \citep{keetha2025mapanything} & 61.21
& 48.53 & 20\% & 79.34
& 70.59 & 11\% & 89.54
& 84.58 & 6\% \\
DepthAnything3 \citep{lin2025depth} & \textbf{89.74}
& 59.85 & 33\% & \textbf{94.87}
& 79.26 & 16\% & \textbf{97.43}
& 89.57 & 8\% \\
\hline
\end{tabular*}
\label{tab:ff_pose}
\end{table*}

\begin{table*}[tb]
\centering
\small
\caption{Quantitative comparisons of the averaged point-prediction performance of feedforward models. Quantitative values are in centimeters. Percentages denote the relative decrease with respect to the corresponding metric on clean views.}
\setlength{\tabcolsep}{0pt}
\setlength{\extrarowheight}{3pt}
\begin{tabular*}{\linewidth}{@{\extracolsep{\fill}}l @{\hspace{4pt}}|ccc ccc ccc ccc}
\hline
\textbf{Methods} &
\multicolumn{3}{c}{Dep.L1$\downarrow$} &
\multicolumn{3}{c}{Acc.$\downarrow$} &
\multicolumn{3}{c}{Comp.$\downarrow$} &
\multicolumn{3}{c}{F1$\uparrow$} \\
\cmidrule(lr){2-4} \cmidrule(lr){5-7} \cmidrule(lr){8-10} \cmidrule(lr){11-13}
\noalign{\vskip -2pt}
 & Clean & Degrad & Err & Clean & Degrad & Err & Clean & Degrad & Err & Clean & Degrad & Err \\
\hline
VGGT \citep{wang2025vggt} & 6.1
& \textbf{13.9} & 128\% & 8.4
& 9.1  & \textbf{8\%}   & 8.4
& 9.2  & \textbf{10\%}  & 22.4
& 14.0 & 38\%  \\
Pi3 \citep{wang2025pi} & 6.5
& 14.3 & 120\% & 7.8
& 8.7  & 12\%  & 7.5
& 8.6  & \textbf{10\%}  & 38.0
& 22.4 & 41\%  \\
MapAnything \citep{keetha2025mapanything} & 16.4
& 27.7 & \textbf{68\%}  & \textbf{4.5}
& \textbf{6.5} & 44\%  & \textbf{4.3}
& \textbf{6.0} & 40\%  & \textbf{78.2}
& \textbf{55.7} & \textbf{29\%} \\
DepthAnything3 \citep{lin2025depth} & \textbf{5.3}
& 15.6 & 194\% & 5.3
& 7.6  & 43\%  & 5.3
& 6.5  & 22\%  & 63.6
& 44.1 & 31\%  \\
\hline
\end{tabular*}
\label{tab:ff_points}
\end{table*}

\subsubsection{Dynamic \& Reflection Occlusion}
Reconstruction in the presence of occluders and dynamic content draws increasing attention, since real-world captures rarely remain static. RealX3D emphasizes evaluating the reconstruction of static scene components under distractors, which is a prerequisite for reliable 3D reconstruction and also provides a stable foundation for 4D modeling \citep{meuleman2025fly,wang2025degauss}.

We categorize occlusions into two representative types: dynamic objects across viewpoints and semi-transparent transient reflections, both of which frequently appear in practical capture. These occlusions can occupy large image regions, exhibit blurred boundaries, and introduce ghosting, thereby breaking multi-view photometric and geometric consistency. We evaluate GS-W \citep{zhang2024gaussian}, Wild3A \citep{li2025wild3a}, SpotlessSplats \citep{sabour2025spotlesssplats}, and DeSplat \citep{wang2025desplat}, which address occlusions from different perspectives, including semantic masking, uncertainty modeling, diffusion priors \citep{rombach2022high}, and transient-field modeling.

Quantitative results in \Cref{tab:photometric_eval} show that SpotlessSplats and DeSplat achieve the strongest performance in dynamic scenes, while improvements under reflection occlusion remain limited across all baselines. In \Cref{fig:render_dynamic}, SpotlessSplats \citep{sabour2025spotlesssplats} and DeSplat \citep{wang2025desplat} can largely remove moving people and foreground clutter using multi-view consistency without predefined object categories. However, failures remain in fine structures near the occlusion boundaries, where the methods tend to leave residual fragments of the dynamic objects and produce texture bleeding into the background, leading to locally distorted geometry and smeared details in the zoomed regions.

Reflection scenes in \Cref{fig:render_reflection} are more challenging. Across all methods, semi-transparent reflections are frequently fused into the static scene as spurious surfaces or faint floating textures, and the transition regions around reflective boundaries show clear ghost trails. These artifacts indicate that current prior-based and transient modeling mechanisms are insufficient to separate semi-transparent occluders from true surface appearance, and the resulting photometric ambiguity propagates to geometry, causing unstable reconstruction near reflective areas.

\subsubsection{Motion \& Defocus Blur}
In RealX3D, we provide both camera motion blur and defocus blur at two severity levels, mild and strong, enabling a systematic evaluation of recent blur-aware 3D reconstruction methods. We evaluate representative rasterization-based deblurring approaches, including DeBlurring-3DGS \citep{zhang2024gaussian}, Deblur-GS \citep{chen2024deblur}, Bad-Gaussians \citep{zhao2024bad}, BAGS \citep{peng2024bags}, and CoCoGaussian \citep{lee2025cocogaussian}, which explicitly model blur during rendering or optimization.

Under the proposed global defocus setting and the physics-based exposure integration for motion blur, quantitative results in \Cref{tab:photometric_eval} show that existing deblurring baselines often perform on par with, or even worse than, vanilla 3DGS \citep{kerbl20233d}. Qualitative results in \Cref{fig:render_defocus} demonstrate that, under strong defocus blur, current methods struggle to recover sharp structures and fine details, leading to oversmoothed appearance and residual blur artifacts. Similarly, \Cref{fig:render_motion} illustrates the strong motion blur case, where DeBlurring-3DGS \citep{lee2024deblurring} and CoCoGaussian \citep{lee2025cocogaussian} achieve partial improvements in visual clarity, but still exhibit noticeable artifacts, texture distortion, and incomplete recovery of high-frequency details.

These results suggest that many existing approaches rely on implicit assumptions about specific blur models or limited blur distributions, and consequently fail to generalize to the diverse and severe blurring patterns encountered in real-world capture. Blur-robust 3D reconstruction therefore remains an underexplored challenge. Moreover, the observed failures further indicate the potential necessity of stronger priors, such as generative or learned appearance models, to handle heavily blur-degraded observations where fine details are largely destroyed \citep{choi2025exploiting,lee2025diet,kong2025rogsplat}.

\begin{figure*}[!tp]
\centering
\includegraphics[width=\textwidth]{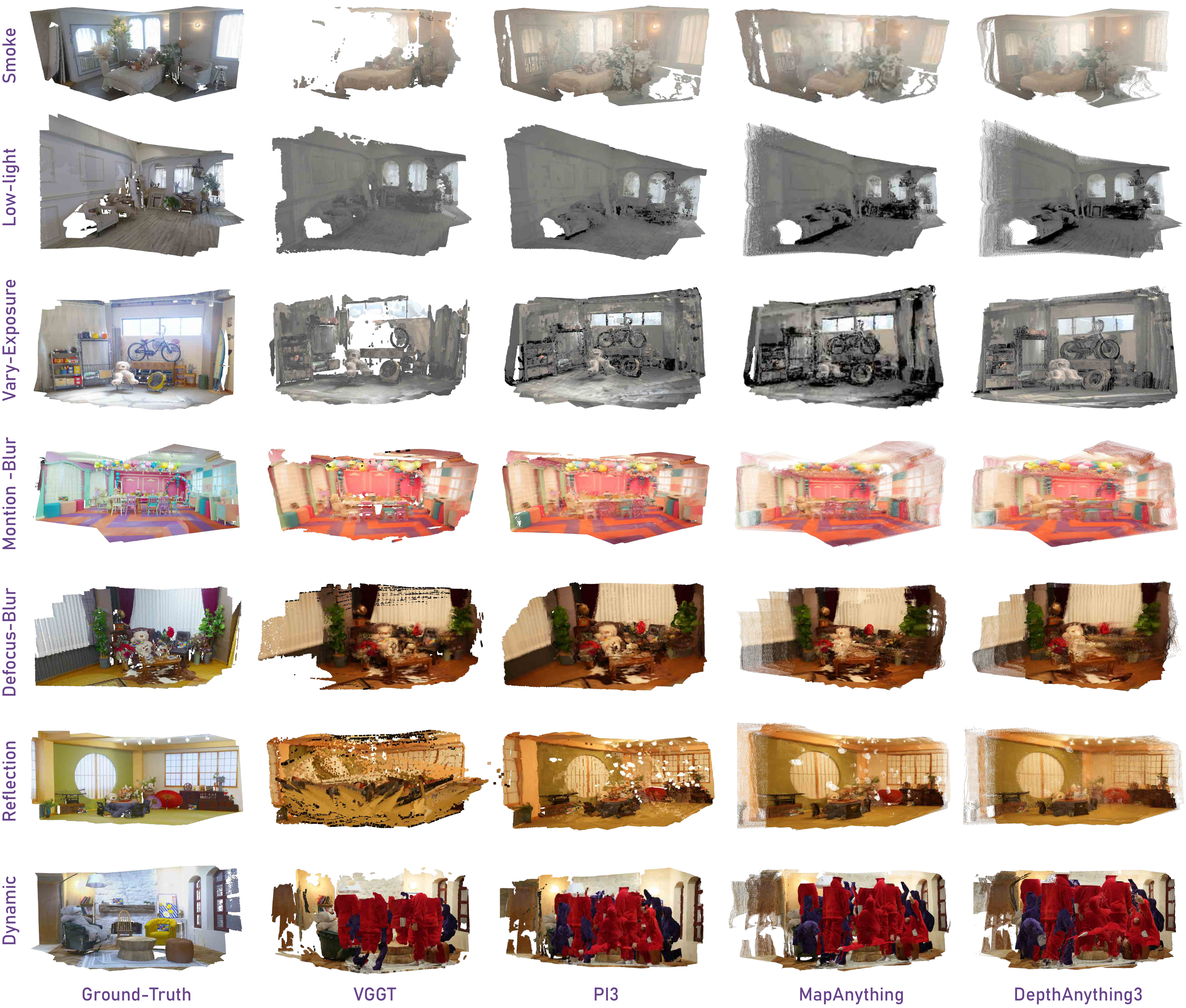}
\caption{Visualizations of point clouds predicted by feed-forward foundation models on smoke, low-light, varying exposure, motion blur, defocus blur, reflection, and dynamic occlusion. For low-light and varying-exposure scenes, the point cloud brightness is adjusted for better visibility.}
\label{fig:point_cloud}
\end{figure*}

\subsection{Geometric Evaluation}
Recent advances in feedforward foundation models have enabled pose-free, zero-shot 3D inference, allowing camera pose estimation and geometry prediction without scene-specific optimization. While these models demonstrate strong performance under ideal conditions, their robustness to real-world degradations remains insufficiently characterized. Leveraging accurate ground-truth camera poses and laser-scanned geometry provided by RealX3D, we systematically benchmark both pose accuracy and geometry quality, reporting results averaged across degradation categories.

\begin{figure*}[t]
  \centering
  \begin{subfigure}[t]{0.49\textwidth}
    \centering
    \includegraphics[width=\linewidth]{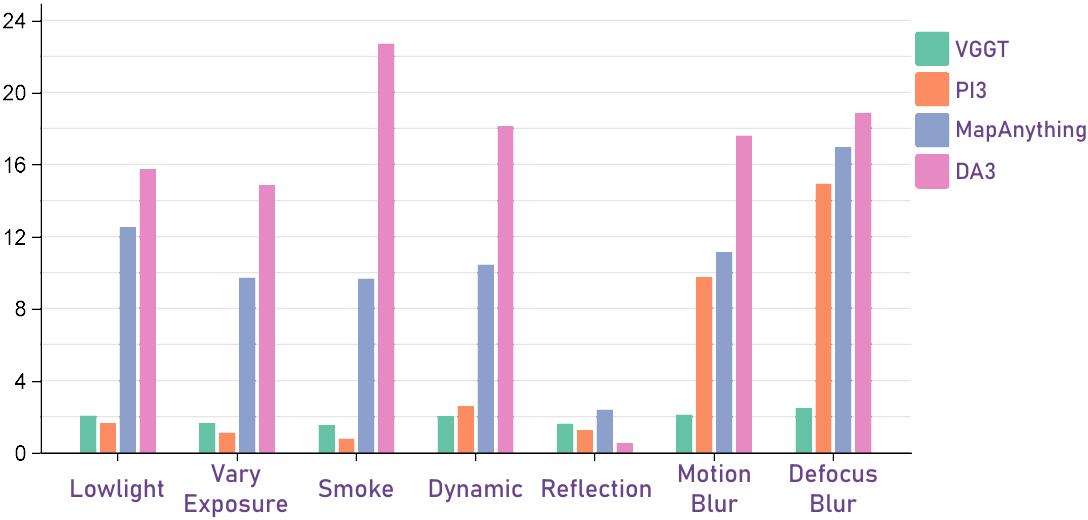}
  \end{subfigure}
  \hfill
  \begin{subfigure}[t]{0.49\textwidth}
    \centering
    \includegraphics[width=\linewidth]{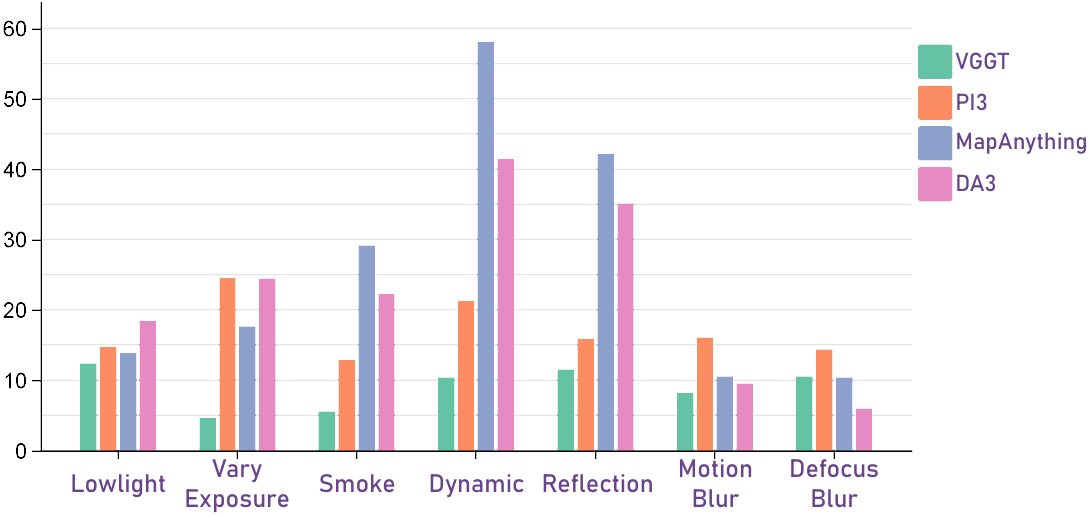}
  \end{subfigure}
  \label{fig:pose_f1_drop}
  \caption{Visualization of the performance gap of foundation models across different degradation types in the RealX3D benchmark. Left: relative pose accuracy degradation, measured as the percentage drop of AUC@10 from clean to degraded views. Right: relative geometry degradation, measured as the percentage drop of point cloud F1 score from clean to degraded views.}
\end{figure*}

\Cref{tab:ff_pose} reports the pose estimation accuracy of feedforward models. Notably, these models remain surprisingly effective under challenging conditions where traditional SfM pipelines \citep{schonberger2016structure} often fail. Among the evaluated methods, DepthAnything3 \citep{lin2025depth} achieves state-of-the-art pose accuracy on clean views; however, it is also the most sensitive to visual degradations, exhibiting the largest relative performance drops under low-quality observations. In contrast, VGGT \citep{wang2025vggt} consistently delivers strong pose accuracy while maintaining the smallest degradation-induced drops, indicating superior robustness across adverse conditions.

\Cref{tab:ff_points} summarizes depth and geometry evaluation results. VGGT \citep{wang2025vggt}, Pi3 \citep{wang2025pi}, and DepthAnything3 \citep{lin2025depth} achieve strong depth prediction accuracy on clean inputs; however, degradation can cause depth errors to increase dramatically, with relative increases of up to nearly 200\%. For point cloud reconstruction, MapAnything \citep{keetha2025mapanything} and DepthAnything3 \citep{lin2025depth} achieve higher accuracy and completeness under clean conditions, yet their performance degrades substantially when exposed to real-world corruption. Overall, VGGT \citep{wang2025vggt} and Pi3 \citep{wang2025pi} exhibit more robust behavior across degradations, whereas MapAnything \citep{keetha2025mapanything} and DepthAnything3 \citep{lin2025depth}, although more accurate in ideal settings, are markedly more sensitive to visual corruption.

\Cref{fig:point_cloud} visualizes representative point cloud predictions under different degradation types. While foundation models generally preserve the global scene layout, they exhibit reduced completeness and loss of fine-grained geometric detail, particularly in regions affected by severe degradation. To further analyze robustness, \Cref{fig:pose_f1_drop} illustrates the relative performance degradation, measured as the percentage drop in pose AUC@10 and point cloud F1 score from clean to degraded views. For pose estimation, blur-related degradations, including motion and defocus blur, induce the largest accuracy drops, as global image blur disrupts keypoint localization and feature correspondence across views. In contrast, point cloud quality is more strongly affected by dynamic occlusions and reflections, which introduce view-inconsistent content that corrupts geometry aggregation. Recent studies on feedforward 4D reconstruction have shown that explicit masking strategies or uncertainty-aware attention can mitigate this dynamic interference \citep{zhang2024monst3r,zhuo2025streaming,chen2025easi3r,feng2025st4rtrack}. Beyond occlusion, illumination variation and scattering also lead to pronounced degradation in geometry quality, indicating that appearance distortions can propagate into cross-view fusion even when the scene remains static. Taken together, these trends highlight a clear divergence in failure modes between pose inference and geometry reconstruction under real-world degradations.

\section{Conclusion}
\label{sec:conclusion}

In this work, we present RealX3D, a real-world benchmark for 3D restoration and reconstruction that covers a broad range of physical degradations across diverse scenes. Unlike previous work that relies on disparate single-degradation datasets, RealX3D provides rich, high-resolution, pixel-aligned image pairs together with accurately scanned point clouds, enabling comprehensive photometric and geometric evaluation. Extensive experiments reveal the advancements and limitations of recent degradation-aware reconstruction and feedforward models, highlighting that robust 3D reconstruction under real-world conditions remains an active challenge.

\backmatter





\bmhead{Conflict of Interest Statement}
The authors declare that they have no conflict of
interest.

\bmhead{Funding Information}
This work was partially supported by JST Moonshot R\&D Grant Number JPMJPS2011, CREST Grant Number JPMJCR2015 and Basic Research Grant (Super AI) of the Institute for AI and Beyond of the University of Tokyo. Shuhong Liu, Xuangeng Chu, Ryo Umagami, and Tomohiro Hashimoto are also supported by JST SPRING, Grant Number JPMJSP2108.

\bmhead{Data Availability}
All data from the RealX3D benchmark will be publicly available in April 2026 under the Creative Commons Attribution 4.0 International License (CC BY 4.0).

\begin{appendices}

\section{Additional Results}
\label{sec:appendix_A}

\Cref{tab:photometric_eval} reports the average performance of diverse baseline models across each degradation category. In this section, we provide per-scene qualitative results for detailed reference. Specifically, \Cref{tab:sup_lowlight}, \cref{tab:sup_exposure}, \Cref{tab:sup_scattering}, \Cref{tab:sup_dynamic}, and \Cref{tab:sup_reflection} summarize training-view restoration and novel-view synthesis results under low-light, smoke scattering, dynamic occlusion, and reflection in RealX3D. \Cref{tab:sup_motion100} and \Cref{tab:sup_motion300} present results for camera motion blur at mild (2cm) and severe (6cm) levels, while \Cref{tab:sup_defocus60} and \Cref{tab:sup_defocus40} present results for defocus blur at mild (0.6m) and severe (0.4m) levels.

In addition, \Cref{tab:sup_ff_pose} and \Cref{tab:sup_ff_point} report the average accuracy of pose estimation and point cloud prediction across degradation types.

In all quantitative tables, the best three results are highlighted as \colorbox{fst}{\textbf{first}}, \colorbox{sec}{\textbf{second}}, and \colorbox{thd}{\textbf{third}}

\begin{sidewaystable}
    \centering
    \caption{Comprehensive per-scene evaluation of photometric fidelity under extreme \textit{\textbf{Low-light}} degradation in RealX3D benchmark.}
    \setlength{\tabcolsep}{1pt}
    \setlength{\extrarowheight}{3pt}
    \scriptsize
\begin{tabularx}{\textwidth}{l @{\hspace{2pt}} | c | *{20}{>{\centering\arraybackslash}X}}
        \Xhline{2\arrayrulewidth}
        \textbf{Methods} & \textbf{Metrics} &
        \multicolumn{2}{c}{BlueHawaii} &
        \multicolumn{2}{c}{Chocolate} &
        \multicolumn{2}{c}{Cupcake} &
        \multicolumn{2}{c}{GearWorks} &
        \multicolumn{2}{c}{Laboratory} &
        \multicolumn{2}{c}{MilkCookie} &
        \multicolumn{2}{c}{Popcorn} &
        \multicolumn{2}{c}{Sculpture} &
        \multicolumn{2}{c}{Ujikintoki} &
        \multicolumn{2}{c}{\textbf{Avg.}} \\
        \hline
        \multirow{3}{*}{\makecell[l]{3DGS\\\citep{kerbl20233d}}} & PSNR$\uparrow$ &\cellcolor{b}7.43&\cellcolor{b}7.49&\cellcolor{b}8.05&\cellcolor{b}7.89&\cellcolor{b}4.39&\cellcolor{b}4.42&\cellcolor{b}6.74&\cellcolor{b}6.90&\cellcolor{b}7.51&\cellcolor{b}7.59&\cellcolor{b}5.77&\cellcolor{b}6.15&\cellcolor{b}7.37&\cellcolor{b}7.46&\cellcolor{b}5.75&\cellcolor{b}5.85&\cellcolor{b}6.24&\cellcolor{b}6.23&\cellcolor{b}6.58&\cellcolor{b}6.66  \\
        & SSIM$\uparrow$ &\cellcolor{b}0.049&\cellcolor{b}0.055&\cellcolor{b}0.089&\cellcolor{b}0.084&\cellcolor{b}0.065&\cellcolor{b}0.064&\cellcolor{b}0.080&\cellcolor{b}0.081&\cellcolor{b}0.039&\cellcolor{b}0.041&\cellcolor{b}0.081&\cellcolor{b}0.052&\cellcolor{b}0.059&\cellcolor{b}0.062&\cellcolor{b}0.019&\cellcolor{b}0.018&\cellcolor{b}0.060&\cellcolor{b}0.062&\cellcolor{b}0.060&\cellcolor{b}0.058  \\
        & LPIPS$\downarrow$ &\cellcolor{thd}\textbf{0.675}&\cellcolor{thd}\textbf{0.671}&\cellcolor{thd}\textbf{0.638}&\cellcolor{thd}\textbf{0.645}&\cellcolor{b}0.623&\cellcolor{b}0.625&\cellcolor{b}0.656&\cellcolor{thd}\textbf{0.658}&\cellcolor{b}0.610&\cellcolor{thd}\textbf{0.610}&\cellcolor{b}0.649&\cellcolor{thd}\textbf{0.653}&\cellcolor{b}0.646&\cellcolor{b}0.650&\cellcolor{b}0.731&\cellcolor{b}0.733&\cellcolor{b}0.681&\cellcolor{b}0.683&\cellcolor{b}0.656&\cellcolor{thd}\textbf{0.659}  \\
\hline
     \end{tabularx}
     \label{tab:sup_lowlight}
\end{sidewaystable}

\begin{sidewaystable}[tb]
    \centering
    \caption{Comprehensive per-scene evaluation of photometric fidelity under extreme \textit{\textbf{Varying Exposure}} degradation in RealX3D benchmark.}
    \setlength{\tabcolsep}{1pt}
    \setlength{\extrarowheight}{4pt}
    \scriptsize
    \begin{tabularx}{\textwidth}{l @{\hspace{2pt}} | c | *{20}{>{\centering\arraybackslash}X}}
        \Xhline{2\arrayrulewidth}
        \textbf{Methods} & \textbf{Metrics} &
        \multicolumn{2}{c}{BlueHawaii} &
        \multicolumn{2}{c}{Chocolate} &
        \multicolumn{2}{c}{Cupcake} &
        \multicolumn{2}{c}{GearWorks} &
        \multicolumn{2}{c}{Laboratory} &
        \multicolumn{2}{c}{MilkCookie} &
        \multicolumn{2}{c}{Popcorn} &
        \multicolumn{2}{c}{Sculpture} &
        \multicolumn{2}{c}{Ujikintoki} &
        \multicolumn{2}{c}{\textbf{Avg.}} \\
        \hline

        \multirow{3}{*}{\makecell[l]{3DGS\\\citep{kerbl20233d}}} & PSNR$\uparrow$ &\cellcolor{thd}\textbf{7.92}&\cellcolor{thd}\textbf{7.90}&\cellcolor{thd}\textbf{8.75}&\cellcolor{thd}\textbf{8.56}&\cellcolor{thd}\textbf{5.30}&\cellcolor{thd}\textbf{5.37}&\cellcolor{thd}\textbf{7.49}&\cellcolor{thd}\textbf{7.80}&\cellcolor{thd}\textbf{7.97}&\cellcolor{thd}\textbf{8.14}&\cellcolor{thd}\textbf{6.90}&\cellcolor{thd}\textbf{7.07}&\cellcolor{thd}\textbf{8.05}&\cellcolor{thd}\textbf{8.12}&\cellcolor{thd}\textbf{6.33}&\cellcolor{thd}\textbf{6.40}&\cellcolor{thd}\textbf{6.92}&\cellcolor{thd}\textbf{7.15}&\cellcolor{thd}\textbf{7.29}&\cellcolor{thd}\textbf{7.39}  \\
        & SSIM$\uparrow$ &\cellcolor{thd}\textbf{0.107}&\cellcolor{thd}\textbf{0.120}&\cellcolor{thd}\textbf{0.149}&\cellcolor{thd}\textbf{0.146}&\cellcolor{thd}\textbf{0.165}&\cellcolor{thd}\textbf{0.161}&\cellcolor{thd}\textbf{0.131}&\cellcolor{thd}\textbf{0.151}&\cellcolor{thd}\textbf{0.093}&\cellcolor{thd}\textbf{0.118}&\cellcolor{thd}\textbf{0.177}&\cellcolor{thd}\textbf{0.161}&\cellcolor{thd}\textbf{0.115}&\cellcolor{thd}\textbf{0.120}&\cellcolor{thd}\textbf{0.048}&\cellcolor{thd}\textbf{0.050}&\cellcolor{thd}\textbf{0.135}&\cellcolor{thd}\textbf{0.155}&\cellcolor{thd}\textbf{0.124}&\cellcolor{thd}\textbf{0.131}  \\
\hline
     \end{tabularx}
     \label{tab:sup_exposure}
\end{sidewaystable}

\begin{sidewaystable}[tb]
    \centering
    \caption{Comprehensive per-scene evaluation of photometric fidelity under \textit{\textbf{Smoke}} degradation in RealX3D benchmark.}
    \setlength{\tabcolsep}{1pt}
    \setlength{\extrarowheight}{3pt}
    \scriptsize
    \begin{tabularx}{\textwidth}{l @{\hspace{2pt}} | c | *{12}{>{\centering\arraybackslash}X}}
        \Xhline{2\arrayrulewidth}
        \textbf{Methods} & \textbf{Metrics} &
        \multicolumn{2}{c}{Akikaze} &
        \multicolumn{2}{c}{Hinoki} &
        \multicolumn{2}{c}{Koharu} &
        \multicolumn{2}{c}{Natsume} &
        \multicolumn{2}{c}{Shirohana} &
        \multicolumn{2}{c}{\textbf{Avg.}} \\
        \hline

        \multirow{3}{*}{\makecell[l]{3DGS\\\citep{kerbl20233d}}} & PSNR$\uparrow$ &\cellcolor{sec}\textbf{10.97}&\cellcolor{sec}\textbf{10.80}&\cellcolor{b}7.53&\cellcolor{b}7.47&\cellcolor{sec}\textbf{11.88}&\cellcolor{sec}\textbf{11.87}&\cellcolor{fst}\textbf{9.96}&\cellcolor{fst}\textbf{9.74}&\cellcolor{sec}\textbf{9.00}&\cellcolor{sec}\textbf{8.91}&\cellcolor{thd}\textbf{9.87}&\cellcolor{thd}\textbf{9.76}  \\
        & SSIM$\uparrow$ &\cellcolor{fst}\textbf{0.573}&\cellcolor{fst}\textbf{0.548}&\cellcolor{b}0.323&\cellcolor{b}0.315&\cellcolor{sec}\textbf{0.634}&\cellcolor{sec}\textbf{0.611}&\cellcolor{fst}\textbf{0.621}&\cellcolor{fst}\textbf{0.604}&\cellcolor{fst}\textbf{0.434}&\cellcolor{fst}\textbf{0.419}&\cellcolor{fst}\textbf{0.517}&\cellcolor{fst}\textbf{0.499}  \\
        & LPIPS$\downarrow$ &\cellcolor{fst}\textbf{0.627}&\cellcolor{fst}\textbf{0.659}&\cellcolor{sec}\textbf{0.669}&\cellcolor{thd}\textbf{0.683}&\cellcolor{fst}\textbf{0.544}&\cellcolor{fst}\textbf{0.573}&\cellcolor{fst}\textbf{0.589}&\cellcolor{sec}\textbf{0.633}&\cellcolor{sec}\textbf{0.716}&\cellcolor{thd}\textbf{0.749}&\cellcolor{fst}\textbf{0.629}&\cellcolor{fst}\textbf{0.659}  \\
        \hline

        \multirow{3}{*}{\makecell[l]{SeaThru-NeRF\\\citep{levy2023seathru}}} & PSNR$\uparrow$ &\cellcolor{b}8.34&\cellcolor{b}8.17&\cellcolor{b}4.89&\cellcolor{b}4.90&\cellcolor{b}7.74&\cellcolor{b}7.77&\cellcolor{thd}\textbf{8.47}&\cellcolor{thd}\textbf{8.48}&\cellcolor{thd}\textbf{8.46}&\cellcolor{thd}\textbf{8.43}&\cellcolor{b}7.58&\cellcolor{b}7.55  \\
\hline
     \end{tabularx}
     \label{tab:sup_scattering}
\end{sidewaystable}

\begin{sidewaystable}[tb]
    \centering
    \caption{Comprehensive per-scene evaluation of photometric fidelity under \textit{\textbf{Dynamic Occlusion}} degradation in RealX3D benchmark.}
    \setlength{\tabcolsep}{1pt}
    \setlength{\extrarowheight}{3pt}
    \scriptsize
    \begin{tabularx}{\textwidth}{l @{\hspace{2pt}} | c | *{18}{>{\centering\arraybackslash}X}}
        \Xhline{2\arrayrulewidth}
        \textbf{Methods} & \textbf{Metrics} &
        \multicolumn{2}{c}{Chocolate} &
        \multicolumn{2}{c}{Cupcake} &
        \multicolumn{2}{c}{GearWorks} &
        \multicolumn{2}{c}{Laboratory} &
        \multicolumn{2}{c}{Limon} &
        \multicolumn{2}{c}{MilkCookie} &
        \multicolumn{2}{c}{Popcorn} &
        \multicolumn{2}{c}{Ujikintoki} &
        \multicolumn{2}{c}{\textbf{Avg.}} \\
        \hline

        \multirow{3}{*}{\makecell[l]{3DGS\\\citep{kerbl20233d}}} & PSNR$\uparrow$ &\cellcolor{thd}\textbf{23.83}&\cellcolor{b}21.41&\cellcolor{b}21.10&\cellcolor{b}17.41&\cellcolor{b}21.92&\cellcolor{b}18.52&\cellcolor{sec}\textbf{21.66}&\cellcolor{thd}\textbf{19.16}&\cellcolor{thd}\textbf{14.20}&\cellcolor{b}13.82&\cellcolor{thd}\textbf{22.49}&\cellcolor{b}19.32&\cellcolor{thd}\textbf{27.61}&\cellcolor{thd}\textbf{25.71}&\cellcolor{thd}\textbf{25.51}&\cellcolor{b}23.25&\cellcolor{thd}\textbf{22.29}&\cellcolor{b}19.83  \\
        & SSIM$\uparrow$ &\cellcolor{fst}\textbf{0.864}&\cellcolor{thd}\textbf{0.761}&\cellcolor{thd}\textbf{0.864}&\cellcolor{b}0.681&\cellcolor{sec}\textbf{0.829}&\cellcolor{thd}\textbf{0.745}&\cellcolor{sec}\textbf{0.851}&\cellcolor{b}0.744&\cellcolor{thd}\textbf{0.568}&\cellcolor{b}0.482&\cellcolor{thd}\textbf{0.887}&\cellcolor{b}0.802&\cellcolor{sec}\textbf{0.892}&\cellcolor{thd}\textbf{0.856}&\cellcolor{thd}\textbf{0.875}&\cellcolor{b}0.837&\cellcolor{thd}\textbf{0.829}&\cellcolor{b}0.739  \\
\hline
     \end{tabularx}
     \label{tab:sup_dynamic}
\end{sidewaystable}

\begin{sidewaystable}[tb]
    \centering
    \caption{Comprehensive per-scene evaluation of photometric fidelity under \textit{\textbf{Reflection}} degradation in RealX3D benchmark.}
    \setlength{\tabcolsep}{1pt}
    \setlength{\extrarowheight}{3pt}
    \scriptsize
    \begin{tabularx}{\textwidth}{l @{\hspace{2pt}} | c | *{18}{>{\centering\arraybackslash}X}}
        \Xhline{2\arrayrulewidth}
        \textbf{Methods} & \textbf{Metrics} &
        \multicolumn{2}{c}{Chocolate} &
        \multicolumn{2}{c}{Cupcake} &
        \multicolumn{2}{c}{GearWorks} &
        \multicolumn{2}{c}{Laboratory} &
        \multicolumn{2}{c}{Limon} &
        \multicolumn{2}{c}{MilkCookie} &
        \multicolumn{2}{c}{Popcorn} &
        \multicolumn{2}{c}{Ujikintoki} &
        \multicolumn{2}{c}{\textbf{Avg.}} \\
        \hline

        \multirow{3}{*}{\makecell[l]{3DGS\\\citep{kerbl20233d}}} & PSNR$\uparrow$ &\cellcolor{b}22.66&\cellcolor{b}21.95&\cellcolor{b}22.98&\cellcolor{b}18.59&\cellcolor{fst}\textbf{26.17}&\cellcolor{sec}\textbf{19.01}&\cellcolor{thd}\textbf{20.38}&\cellcolor{b}19.47&\cellcolor{sec}\textbf{23.85}&\cellcolor{thd}\textbf{22.00}&\cellcolor{thd}\textbf{23.04}&\cellcolor{b}20.99&\cellcolor{fst}\textbf{26.51}&\cellcolor{fst}\textbf{25.79}&\cellcolor{thd}\textbf{26.97}&\cellcolor{b}25.00&\cellcolor{thd}\textbf{24.07}&\cellcolor{b}21.60  \\
        & SSIM$\uparrow$ &\cellcolor{fst}\textbf{0.792}&\cellcolor{sec}\textbf{0.752}&\cellcolor{thd}\textbf{0.885}&\cellcolor{thd}\textbf{0.741}&\cellcolor{fst}\textbf{0.864}&\cellcolor{sec}\textbf{0.789}&\cellcolor{sec}\textbf{0.818}&\cellcolor{b}0.712&\cellcolor{thd}\textbf{0.715}&\cellcolor{b}0.656&\cellcolor{sec}\textbf{0.896}&\cellcolor{b}0.844&\cellcolor{fst}\textbf{0.879}&\cellcolor{fst}\textbf{0.859}&\cellcolor{fst}\textbf{0.881}&\cellcolor{thd}\textbf{0.858}&\cellcolor{thd}\textbf{0.841}&\cellcolor{thd}\textbf{0.776}  \\
\hline
     \end{tabularx}
     \label{tab:sup_reflection}
\end{sidewaystable}

\begin{sidewaystable}[tb]
    \centering
    \caption{Comprehensive per-scene evaluation of photometric fidelity under \textit{\textbf{mild Motion Blur}} degradation in RealX3D benchmark.}
    \setlength{\tabcolsep}{1pt}
    \setlength{\extrarowheight}{2pt}
    \scriptsize
    \begin{tabularx}{\textwidth}{l @{\hspace{2pt}} | c | *{18}{>{\centering\arraybackslash}X}}
        \Xhline{2\arrayrulewidth}
        \textbf{Methods} & \textbf{Metrics} &
        \multicolumn{2}{c}{Chocolate} &
        \multicolumn{2}{c}{Cupcake} &
        \multicolumn{2}{c}{GearWorks} &
        \multicolumn{2}{c}{Laboratory} &
        \multicolumn{2}{c}{Limon} &
        \multicolumn{2}{c}{MilkCookie} &
        \multicolumn{2}{c}{Popcorn} &
        \multicolumn{2}{c}{Ujikintoki} &
        \multicolumn{2}{c}{\textbf{Avg.}} \\
        \hline

        \multirow{3}{*}{\makecell[l]{3DGS\\\citep{kerbl20233d}}} & PSNR$\uparrow$ &\cellcolor{sec}\textbf{23.19}&\cellcolor{sec}\textbf{22.87}&\cellcolor{fst}\textbf{22.91}&\cellcolor{sec}\textbf{20.32}&\cellcolor{fst}\textbf{22.94}&\cellcolor{fst}\textbf{18.15}&\cellcolor{sec}\textbf{22.00}&\cellcolor{sec}\textbf{21.68}&\cellcolor{fst}\textbf{22.99}&\cellcolor{fst}\textbf{21.91}&\cellcolor{fst}\textbf{24.26}&\cellcolor{sec}\textbf{21.28}&\cellcolor{sec}\textbf{22.11}&\cellcolor{sec}\textbf{21.91}&\cellcolor{fst}\textbf{24.78}&\cellcolor{fst}\textbf{24.43}&\cellcolor{fst}\textbf{23.15}&\cellcolor{fst}\textbf{21.57}  \\
        & SSIM$\uparrow$ &\cellcolor{sec}\textbf{0.695}&\cellcolor{sec}\textbf{0.675}&\cellcolor{sec}\textbf{0.765}&\cellcolor{thd}\textbf{0.694}&\cellcolor{sec}\textbf{0.739}&\cellcolor{sec}\textbf{0.694}&\cellcolor{sec}\textbf{0.700}&\cellcolor{sec}\textbf{0.680}&\cellcolor{sec}\textbf{0.715}&\cellcolor{sec}\textbf{0.680}&\cellcolor{fst}\textbf{0.798}&\cellcolor{sec}\textbf{0.744}&\cellcolor{sec}\textbf{0.703}&\cellcolor{sec}\textbf{0.693}&\cellcolor{fst}\textbf{0.772}&\cellcolor{fst}\textbf{0.765}&\cellcolor{sec}\textbf{0.736}&\cellcolor{sec}\textbf{0.703}  \\
\hline
     \end{tabularx}
     \label{tab:sup_motion100}
\end{sidewaystable}

\begin{sidewaystable}[tb]
    \centering
    \caption{Comprehensive per-scene evaluation of photometric fidelity under \textit{\textbf{strong Motion Blur}} degradation in RealX3D benchmark.}
    \setlength{\tabcolsep}{2pt}
    \setlength{\extrarowheight}{2pt}
    \scriptsize
    \begin{tabularx}{\textwidth}{l @{\hspace{2pt}} | c | *{18}{>{\centering\arraybackslash}X}}
        \Xhline{2\arrayrulewidth}
        \textbf{Methods} & \textbf{Metrics} &
        \multicolumn{2}{c}{Chocolate} &
        \multicolumn{2}{c}{Cupcake} &
        \multicolumn{2}{c}{GearWorks} &
        \multicolumn{2}{c}{Laboratory} &
        \multicolumn{2}{c}{Limon} &
        \multicolumn{2}{c}{MilkCookie} &
        \multicolumn{2}{c}{Popcorn} &
        \multicolumn{2}{c}{Ujikintoki} &
        \multicolumn{2}{c}{\textbf{Avg.}} \\
        \hline

        \multirow{3}{*}{\makecell[l]{3DGS\\\citep{kerbl20233d}}} & PSNR$\uparrow$ &\cellcolor{sec}\textbf{20.05}&\cellcolor{thd}\textbf{19.88}&\cellcolor{sec}\textbf{20.30}&\cellcolor{b}18.38&\cellcolor{fst}\textbf{20.07}&\cellcolor{thd}\textbf{17.13}&\cellcolor{thd}\textbf{19.16}&\cellcolor{thd}\textbf{18.91}&\cellcolor{sec}\textbf{19.99}&\cellcolor{sec}\textbf{19.43}&\cellcolor{fst}\textbf{21.42}&\cellcolor{thd}\textbf{20.18}&\cellcolor{thd}\textbf{19.58}&\cellcolor{thd}\textbf{19.53}&\cellcolor{thd}\textbf{22.05}&\cellcolor{thd}\textbf{21.86}&\cellcolor{sec}\textbf{20.33}&\cellcolor{thd}\textbf{19.41}  \\
        & SSIM$\uparrow$ &\cellcolor{sec}\textbf{0.596}&\cellcolor{sec}\textbf{0.580}&\cellcolor{sec}\textbf{0.701}&\cellcolor{thd}\textbf{0.640}&\cellcolor{thd}\textbf{0.669}&\cellcolor{thd}\textbf{0.634}&\cellcolor{sec}\textbf{0.628}&\cellcolor{thd}\textbf{0.611}&\cellcolor{sec}\textbf{0.620}&\cellcolor{sec}\textbf{0.591}&\cellcolor{sec}\textbf{0.742}&\cellcolor{thd}\textbf{0.700}&\cellcolor{sec}\textbf{0.630}&\cellcolor{sec}\textbf{0.625}&\cellcolor{fst}\textbf{0.718}&\cellcolor{fst}\textbf{0.714}&\cellcolor{sec}\textbf{0.663}&\cellcolor{sec}\textbf{0.637}  \\
\hline
     \end{tabularx}
     \label{tab:sup_motion300}
\end{sidewaystable}

\begin{sidewaystable}[tb]
    \centering
    \caption{Comprehensive per-scene evaluation of photometric fidelity under \textit{\textbf{mild Defocus Blur}} degradation in RealX3D benchmark.}
    \setlength{\tabcolsep}{1pt}
    \setlength{\extrarowheight}{1.5pt}
    \scriptsize
    \begin{tabularx}{\textwidth}{l @{\hspace{2pt}} | c | *{18}{>{\centering\arraybackslash}X}}
        \Xhline{2\arrayrulewidth}
        \textbf{Methods} & \textbf{Metrics} &
        \multicolumn{2}{c}{Chocolate} &
        \multicolumn{2}{c}{Cupcake} &
        \multicolumn{2}{c}{GearWorks} &
        \multicolumn{2}{c}{Laboratory} &
        \multicolumn{2}{c}{Limon} &
        \multicolumn{2}{c}{MilkCookie} &
        \multicolumn{2}{c}{Popcorn} &
        \multicolumn{2}{c}{Ujikintoki} &
        \multicolumn{2}{c}{\textbf{Avg.}} \\
        \hline

        \multirow{3}{*}{\makecell[l]{3DGS\\\citep{kerbl20233d}}} & PSNR$\uparrow$ &\cellcolor{fst}\textbf{24.68}&\cellcolor{fst}\textbf{24.38}&\cellcolor{sec}\textbf{23.67}&\cellcolor{b}19.99&\cellcolor{sec}\textbf{21.13}&\cellcolor{thd}\textbf{17.57}&\cellcolor{fst}\textbf{21.96}&\cellcolor{thd}\textbf{21.77}&\cellcolor{sec}\textbf{22.37}&\cellcolor{thd}\textbf{21.37}&\cellcolor{fst}\textbf{23.54}&\cellcolor{fst}\textbf{22.12}&\cellcolor{sec}\textbf{21.83}&\cellcolor{sec}\textbf{21.86}&\cellcolor{sec}\textbf{25.29}&\cellcolor{b}25.00&\cellcolor{fst}\textbf{23.06}&\cellcolor{sec}\textbf{21.76}  \\
        & SSIM$\uparrow$ &\cellcolor{sec}\textbf{0.692}&\cellcolor{fst}\textbf{0.677}&\cellcolor{sec}\textbf{0.741}&\cellcolor{thd}\textbf{0.680}&\cellcolor{fst}\textbf{0.687}&\cellcolor{sec}\textbf{0.647}&\cellcolor{fst}\textbf{0.656}&\cellcolor{sec}\textbf{0.647}&\cellcolor{sec}\textbf{0.605}&\cellcolor{thd}\textbf{0.588}&\cellcolor{fst}\textbf{0.749}&\cellcolor{sec}\textbf{0.727}&\cellcolor{sec}\textbf{0.663}&\cellcolor{sec}\textbf{0.659}&\cellcolor{thd}\textbf{0.733}&\cellcolor{b}0.732&\cellcolor{fst}\textbf{0.691}&\cellcolor{fst}\textbf{0.670}  \\
\hline
     \end{tabularx}
     \label{tab:sup_defocus60}
\end{sidewaystable}

\begin{sidewaystable}[tb]
    \centering
    \caption{Comprehensive per-scene evaluation of photometric fidelity under \textit{\textbf{strong Defocus Blur}} degradation in RealX3D benchmark.}
    \setlength{\tabcolsep}{1pt}
    \setlength{\extrarowheight}{1.5pt}
    \scriptsize
    \begin{tabularx}{\textwidth}{l @{\hspace{2pt}} | c | *{18}{>{\centering\arraybackslash}X}}
        \Xhline{2\arrayrulewidth}
        \textbf{Methods} & \textbf{Metrics} &
        \multicolumn{2}{c}{Chocolate} &
        \multicolumn{2}{c}{Cupcake} &
        \multicolumn{2}{c}{GearWorks} &
        \multicolumn{2}{c}{Laboratory} &
        \multicolumn{2}{c}{Limon} &
        \multicolumn{2}{c}{MilkCookie} &
        \multicolumn{2}{c}{Popcorn} &
        \multicolumn{2}{c}{Ujikintoki} &
        \multicolumn{2}{c}{\textbf{Avg.}} \\
        \hline

        \multirow{3}{*}{\makecell[l]{3DGS\\\citep{kerbl20233d}}} & PSNR$\uparrow$ &\cellcolor{sec}\textbf{21.95}&\cellcolor{sec}\textbf{21.80}&\cellcolor{thd}\textbf{21.10}&\cellcolor{b}18.31&\cellcolor{thd}\textbf{18.98}&\cellcolor{thd}\textbf{16.67}&\cellcolor{thd}\textbf{19.76}&\cellcolor{thd}\textbf{19.57}&\cellcolor{thd}\textbf{20.34}&\cellcolor{thd}\textbf{19.76}&\cellcolor{sec}\textbf{21.66}&\cellcolor{b}19.35&\cellcolor{b}19.79&\cellcolor{b}19.92&\cellcolor{b}23.03&\cellcolor{b}22.97&\cellcolor{sec}\textbf{20.83}&\cellcolor{thd}\textbf{19.79}  \\
        & SSIM$\uparrow$ &\cellcolor{sec}\textbf{0.595}&\cellcolor{sec}\textbf{0.585}&\cellcolor{thd}\textbf{0.668}&\cellcolor{b}0.624&\cellcolor{fst}\textbf{0.634}&\cellcolor{sec}\textbf{0.604}&\cellcolor{thd}\textbf{0.599}&\cellcolor{thd}\textbf{0.595}&\cellcolor{sec}\textbf{0.553}&\cellcolor{b}0.542&\cellcolor{fst}\textbf{0.708}&\cellcolor{thd}\textbf{0.682}&\cellcolor{thd}\textbf{0.601}&\cellcolor{thd}\textbf{0.600}&\cellcolor{thd}\textbf{0.687}&\cellcolor{thd}\textbf{0.692}&\cellcolor{fst}\textbf{0.631}&\cellcolor{fst}\textbf{0.616}  \\
\hline
     \end{tabularx}
     \label{tab:sup_defocus40}
\end{sidewaystable}

\begin{sidewaystable}[t]
\centering
\caption{Comprehensive evaluation of pose estimation accuracy under diverse degradation types in RealX3D benchmark.}
\small
\setlength{\tabcolsep}{0pt}
\setlength{\extrarowheight}{2pt}
\begin{tabularx}{\textwidth}{l @{\hspace{2pt}} | @{\hspace{2pt}} *{3}{>{\centering\arraybackslash}X} | *{3}{>{\centering\arraybackslash}X} | *{3}{>{\centering\arraybackslash}X} | *{3}{>{\centering\arraybackslash}X}}
\toprule
& \multicolumn{3}{c}{VGGT \citep{wang2025vggt}} & \multicolumn{3}{c}{Pi3 \citep{wang2025pi}} & \multicolumn{3}{c}{MapAnything \citep{keetha2025mapanything}} & \multicolumn{3}{c}{DepthAnyth.3 \citep{lin2025depth}} \\ 
\cmidrule(lr){2-4} \cmidrule(lr){5-7} \cmidrule(lr){8-10} \cmidrule(lr){11-13}
 Scene & AUC@5 & AUC@10 & AUC@20 & AUC@5 & AUC@10 & AUC@20 & AUC@5 & AUC@10 & AUC@20 & AUC@5 & AUC@10 & AUC@20 \\ 
\midrule
 Clean &\cellcolor{thd}\textbf{86.13}&\cellcolor{thd}\textbf{92.98}&\cellcolor{thd}\textbf{96.49}&\cellcolor{sec}\textbf{86.58}&\cellcolor{sec}\textbf{93.28}&\cellcolor{sec}\textbf{96.64}&\cellcolor{b}61.21&\cellcolor{b}79.34&\cellcolor{b}89.54&\cellcolor{fst}\textbf{89.74}&\cellcolor{fst}\textbf{94.87}&\cellcolor{fst}\textbf{97.43}  \\
 \midrule
 Lowlight &\cellcolor{sec}\textbf{81.85}&\cellcolor{sec}\textbf{90.92}&\cellcolor{sec}\textbf{95.46}&\cellcolor{fst}\textbf{83.33}&\cellcolor{fst}\textbf{91.64}&\cellcolor{fst}\textbf{95.82}&\cellcolor{b}42.39&\cellcolor{b}66.81&\cellcolor{b}82.58&\cellcolor{thd}\textbf{59.12}&\cellcolor{thd}\textbf{79.16}&\cellcolor{thd}\textbf{89.50}  \\
 Vary Exposure &\cellcolor{sec}\textbf{82.86}&\cellcolor{sec}\textbf{91.34}&\cellcolor{sec}\textbf{95.67}&\cellcolor{fst}\textbf{84.42}&\cellcolor{fst}\textbf{92.18}&\cellcolor{fst}\textbf{96.09}&\cellcolor{b}47.63&\cellcolor{b}69.64&\cellcolor{b}83.72&\cellcolor{thd}\textbf{60.68}&\cellcolor{thd}\textbf{80.00}&\cellcolor{thd}\textbf{89.95}  \\
\bottomrule
\end{tabularx}
\label{tab:sup_ff_pose}
\end{sidewaystable}

\begin{sidewaystable}[t]
\centering
\caption{Comprehensive evaluation of depth and point cloud prediction accuracy under diverse degradation types in RealX3D benchmark.}
\small
\setlength{\tabcolsep}{0pt}
\setlength{\extrarowheight}{2pt}
\begin{tabularx}{\textwidth}{l @{\hspace{2pt}} | @{\hspace{2pt}} *{5}{>{\centering\arraybackslash}X} | *{5}{>{\centering\arraybackslash}X} | *{5}{>{\centering\arraybackslash}X} | *{5}{>{\centering\arraybackslash}X}}
\toprule
& \multicolumn{5}{c}{VGGT \citep{wang2025vggt}} & \multicolumn{5}{c}{Pi3 \citep{wang2025pi}} & \multicolumn{5}{c}{MapAnything \citep{keetha2025mapanything}} & \multicolumn{5}{c}{DepthAnyth.3 \citep{lin2025depth}} \\ 
\cmidrule(lr){2-6} \cmidrule(lr){7-11} \cmidrule(lr){12-16} \cmidrule(lr){17-21}
 Scene & Dep$\downarrow$ & Acc$\downarrow$ & Com$\downarrow$ & Over$\downarrow$ & F1\%$\uparrow$ & Dep$\downarrow$ & Acc$\downarrow$ & Com$\downarrow$ & Over$\downarrow$ & F1\%$\uparrow$ & Dep$\downarrow$ & Acc$\downarrow$ & Com$\downarrow$ & Over$\downarrow$ & F1\%$\uparrow$ & Dep$\downarrow$ & Acc$\downarrow$ & Com$\downarrow$ & Over$\downarrow$ & F1\%$\uparrow$ \\ 
 \midrule
 Clean &\cellcolor{sec}\textbf{6.1}&\cellcolor{b}8.4&\cellcolor{b}8.4&\cellcolor{b}8.4&\cellcolor{b}22.4&\cellcolor{thd}\textbf{6.5}&\cellcolor{thd}\textbf{7.8}&\cellcolor{thd}\textbf{7.5}&\cellcolor{thd}\textbf{7.7}&\cellcolor{thd}\textbf{38.0}&\cellcolor{b}16.4&\cellcolor{fst}\textbf{4.5}&\cellcolor{fst}\textbf{4.3}&\cellcolor{fst}\textbf{4.4}&\cellcolor{fst}\textbf{78.2}&\cellcolor{fst}\textbf{5.3}&\cellcolor{sec}\textbf{5.3}&\cellcolor{sec}\textbf{5.3}&\cellcolor{sec}\textbf{5.3}&\cellcolor{sec}\textbf{63.6}  \\
 \midrule
 Lowlight &\cellcolor{fst}\textbf{14.9}&\cellcolor{b}9.2&\cellcolor{b}9.5&\cellcolor{b}9.3&\cellcolor{b}10.1&\cellcolor{sec}\textbf{20.6}&\cellcolor{thd}\textbf{8.6}&\cellcolor{thd}\textbf{8.5}&\cellcolor{thd}\textbf{8.5}&\cellcolor{thd}\textbf{23.3}&\cellcolor{b}44.8&\cellcolor{fst}\textbf{5.9}&\cellcolor{fst}\textbf{5.3}&\cellcolor{fst}\textbf{5.6}&\cellcolor{fst}\textbf{64.4}&\cellcolor{thd}\textbf{25.8}&\cellcolor{sec}\textbf{7.5}&\cellcolor{sec}\textbf{6.3}&\cellcolor{sec}\textbf{6.9}&\cellcolor{sec}\textbf{45.2}  \\
\bottomrule
\end{tabularx}
\label{tab:sup_ff_point}
\end{sidewaystable}




\end{appendices}


\clearpage
\bibliography{sn-bibliography}

\end{document}